\documentclass[10pt,journal,compsoc]{IEEEtran}
\IEEEoverridecommandlockouts
\ifCLASSOPTIONcompsoc
  \usepackage[nocompress]{cite}
\else
  \usepackage{cite}
\fi
\usepackage[normalem]{ulem}
\usepackage{framed,multirow}
\usepackage{amssymb}
\usepackage{latexsym}
\usepackage{url}
\usepackage{xcolor}
\definecolor{newcolor}{rgb}{.8,.349,.1}
\usepackage{soul}
\usepackage{caption}
\usepackage{diagbox}
\usepackage{epsfig}
\usepackage{ragged2e}

\usepackage{amssymb}
\usepackage{amsmath}
\usepackage{amsthm}
\usepackage{bm}
\usepackage{amsfonts}
\usepackage{cases}
\usepackage{esint}
\usepackage{mathtools} 
\usepackage{graphicx}
\usepackage{subfig}
\usepackage{algorithm} 
\usepackage{algorithmic} 
\usepackage{multirow} 
\usepackage{amsmath}
\usepackage{xcolor}

\newcommand{\highlighted}[1]{\emph{#1}}
\newcommand{\cusparagraph}[1]{\noindent\emph{#1}}

\usepackage{tabularx}
\usepackage{makecell}
\usepackage{overpic}
\usepackage{colortbl}
\usepackage{multirow}
\usepackage{multicol}
\usepackage{ulem}

\newcommand{\secref}[1]{Sec.~\ref{#1}}

\newcommand{\figref}[1]{Fig.~\ref{#1}} 

\newcommand{\tabref}[1]{Table~\ref{#1}}
\newcommand{\algref}[1]{Algorithm~\ref{#1}}
\newcommand{\sArt}[0]{state-of-the-art}

\newcommand{\ours}[0]{ESB}

\newcommand{\alexnetOimp}[0]{\highlighted{4.78\%}}
\newcommand{\resnetOimp}[0]{\highlighted{1.92\%}}
\newcommand{\mobileOimp}[0]{\highlighted{3.56\%}}

\newcommand{\papertitle}[0]{
Elastic Significant Bit Quantization and Acceleration for Deep Neural Networks}

\begin{document}
\title{\papertitle}
\author{Cheng~Gong,
        Ye~Lu,
        Kunpeng~Xie,
        Zongming~Jin,
        Tao~Li,
        Yanzhi Wang,~\IEEEmembership{Member,~IEEE}
\IEEEcompsocitemizethanks{
\IEEEcompsocthanksitem Cheng Gong, Kunpeng~Xie and Zongming~Jin are affiliated with the College of Computer Science, Nankai University, Tianjin, China, and the Tianjin Key Laboratory of Network and Data Security Technology.
E-mail: \{cheng-gong,xkp,zongming\_jin\}@mail.nankai.edu.cn.
\IEEEcompsocthanksitem Ye Lu and Tao Li are with the College of Computer Science, Nankai University, Tianjin, China, the Tianjin Key Laboratory of Network and Data Security Technology, and the State Key Laboratory of Computer Architecture (ICT,CAS).\protect
~E-mail: \{luye,litao\}@nankai.edu.cn.
\IEEEcompsocthanksitem Yanzhi Wang is with the Department of Electrical and Computer Engineering, and Khoury College of Computer Science (Affiliated) at Northeastern University.
E-mail: yanz.wang@northeastern.edu.
}
\thanks{Manuscript received Aug. 30, 2021; revised Oct. 30, 2021; accepted Nov. 12, 2021.}
\thanks{(Cheng Gong and Ye Lu contributed equally; they are both first authors, corresponding~authors:~Tao~Li.)}
}
\markboth{IEEE Transactions on Parallel and Distributed Systems}%
{Gong \MakeLowercase{\textit{et al.}}: \papertitle}

\IEEEtitleabstractindextext{
\begin{abstract}
\justifying  
Quantization has been proven to be a vital method for improving the inference efficiency of deep neural networks (DNNs).
However, it is still challenging to strike a good balance between accuracy and efficiency while quantizing DNN weights or activation values from high-precision formats to their quantized counterparts.
We propose a new method called elastic significant bit quantization (\ours) that controls the number of significant bits of quantized values to obtain better inference accuracy with fewer resources.
We design a unified mathematical formula to constrain the quantized values of the \ours~with a flexible number of significant bits. 
We also introduce a distribution difference aligner (DDA) to quantitatively align the distributions between the full-precision weight or activation values and quantized values.
Consequently, \ours~is suitable for various bell-shaped distributions of weights and activation of DNNs, thus maintaining a high inference accuracy. 
Benefitting from fewer significant bits of quantized values, ESB can reduce the multiplication complexity.
We implement \ours~as an accelerator and quantitatively evaluate its efficiency on FPGAs. 
Extensive experimental results illustrate that \ours~quantization consistently outperforms state-of-the-art methods and achieves average accuracy improvements of \alexnetOimp, \resnetOimp, and \mobileOimp~over AlexNet, ResNet18, and MobileNetV2, respectively.
Furthermore, \ours~as an accelerator can achieve \highlighted{10.95 GOPS} peak performance of \highlighted{1k LUTs} without DSPs on the Xilinx ZCU102 FPGA platform.
Compared with CPU, GPU, and \sArt~accelerators on FPGAs, the \ours~accelerator can improve the energy efficiency by up to \highlighted{65$\times$}, \highlighted{11$\times$}, and \highlighted{26$\times$}, respectively.
\end{abstract}
}
\maketitle
\begin{figure*}[!t]
\centering
    \subfloat[Quantized values of PoT methods]{
    \label{fig:dis_of_pot_quantized_values}
    \includegraphics[width=0.49\textwidth]{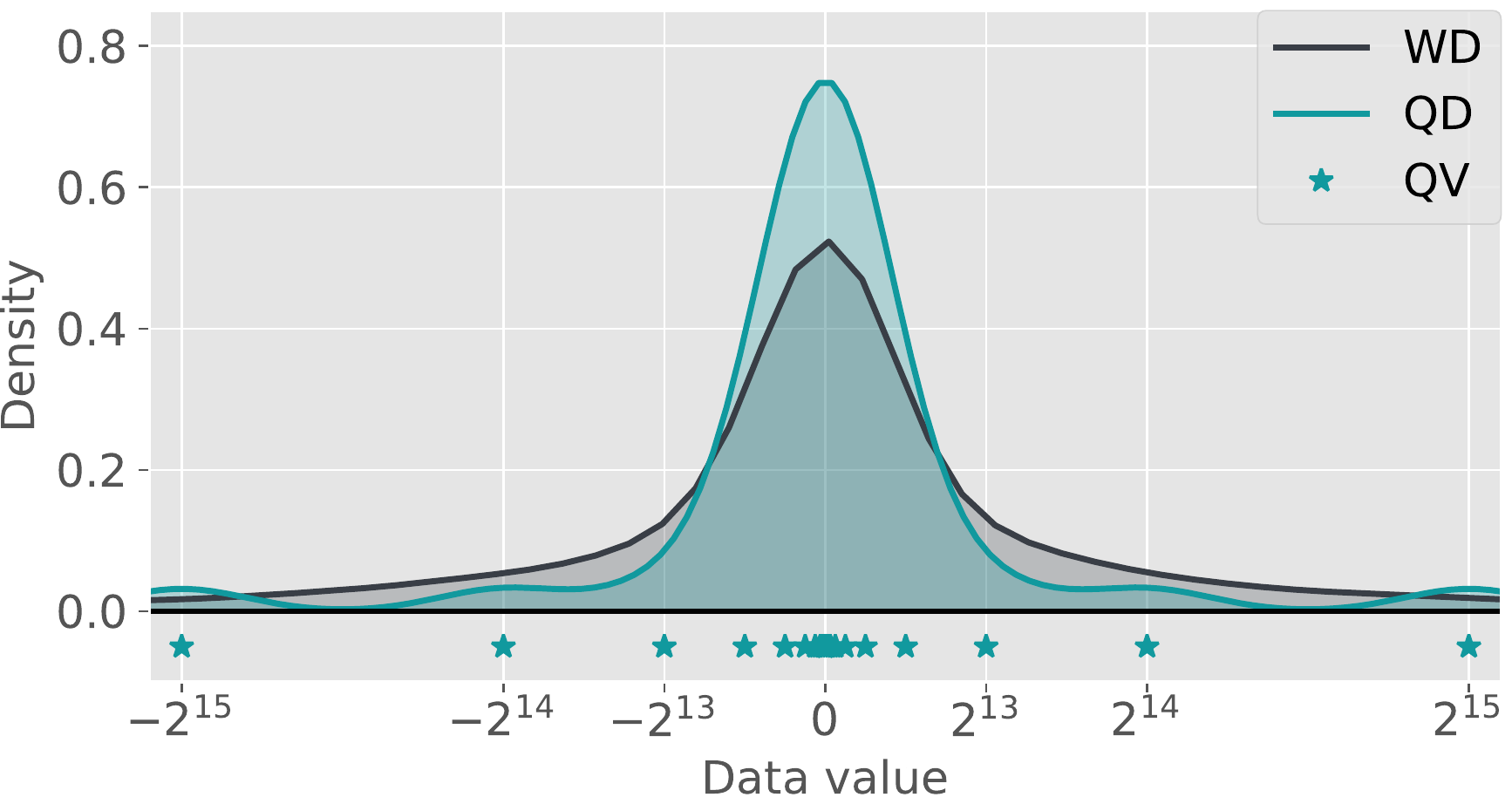}
    }\subfloat[Quantized values of fixed-point methods]{
    \label{fig:dis_of_linearq_quantized_values}
    \includegraphics[width=0.49\textwidth]{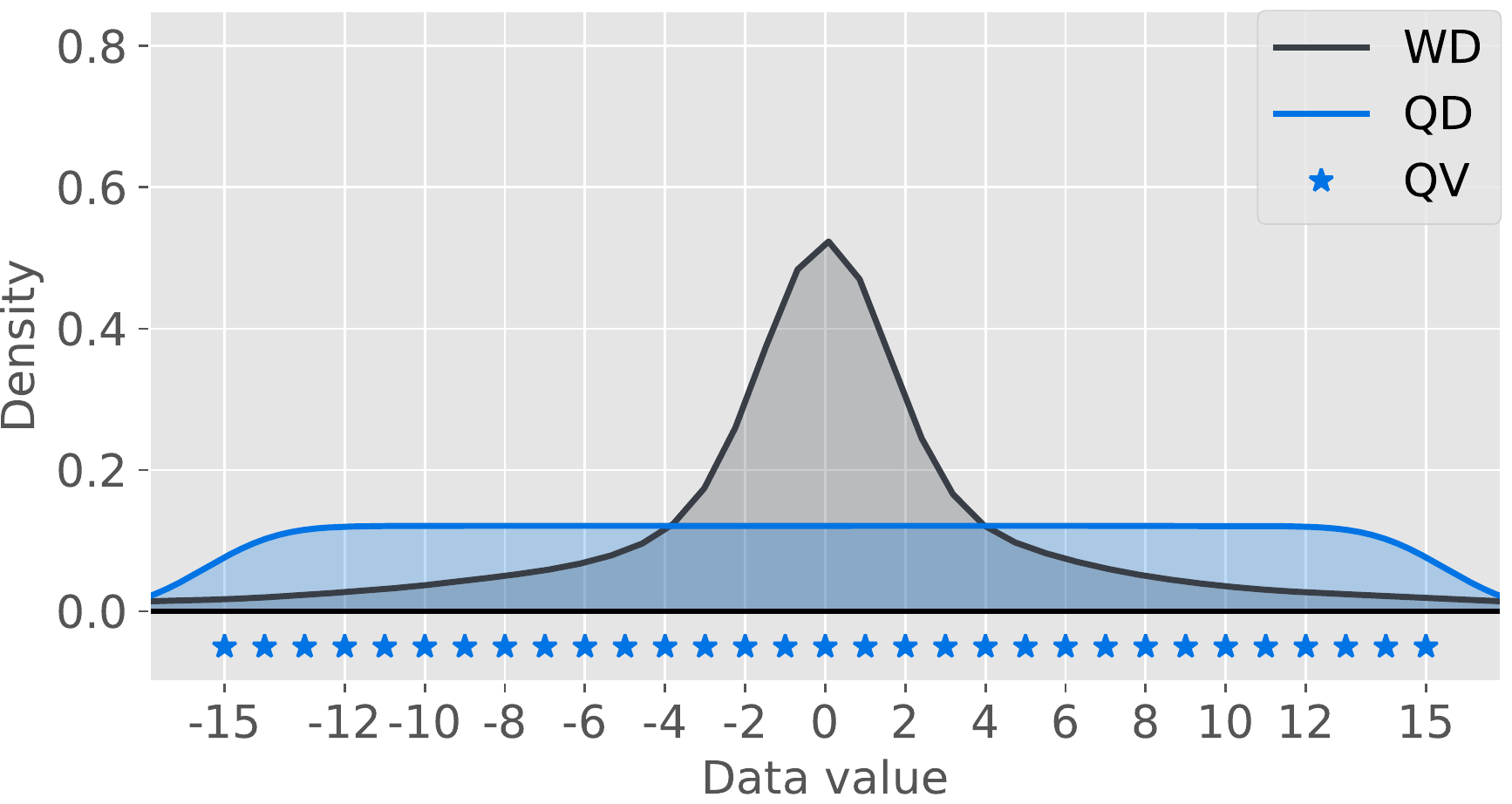}
    }\\\vspace{-10pt}
    \subfloat[Quantized values with 2 significant bits]{
    \label{fig:dis_of_2sb_quantized_values}
    \includegraphics[width=0.49\textwidth]{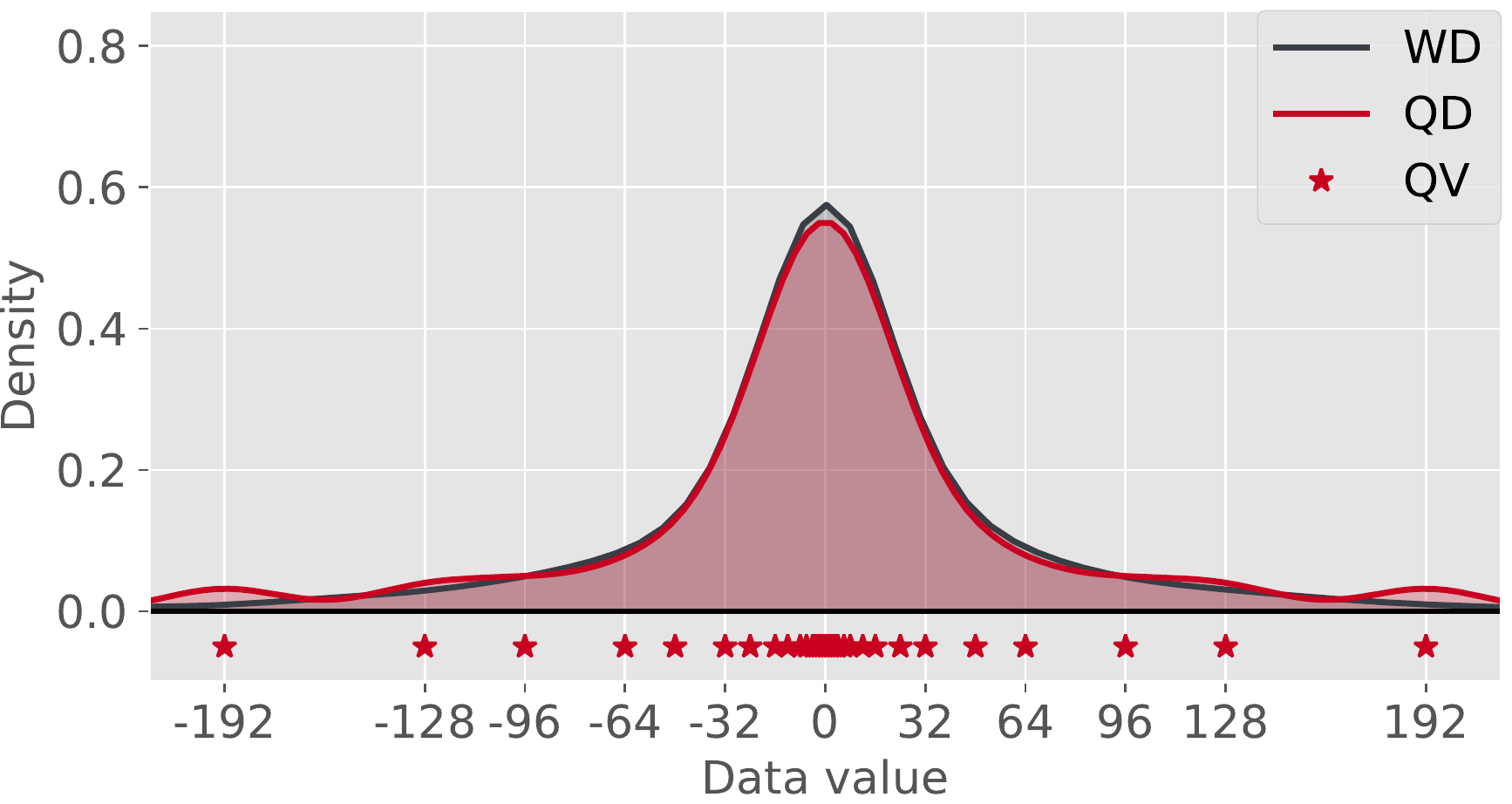}
    }
    \subfloat[Quantized values with elastic significant bits]{
    \label{fig:dis_of_esb_quantized_values}
    \includegraphics[width=0.49\textwidth]{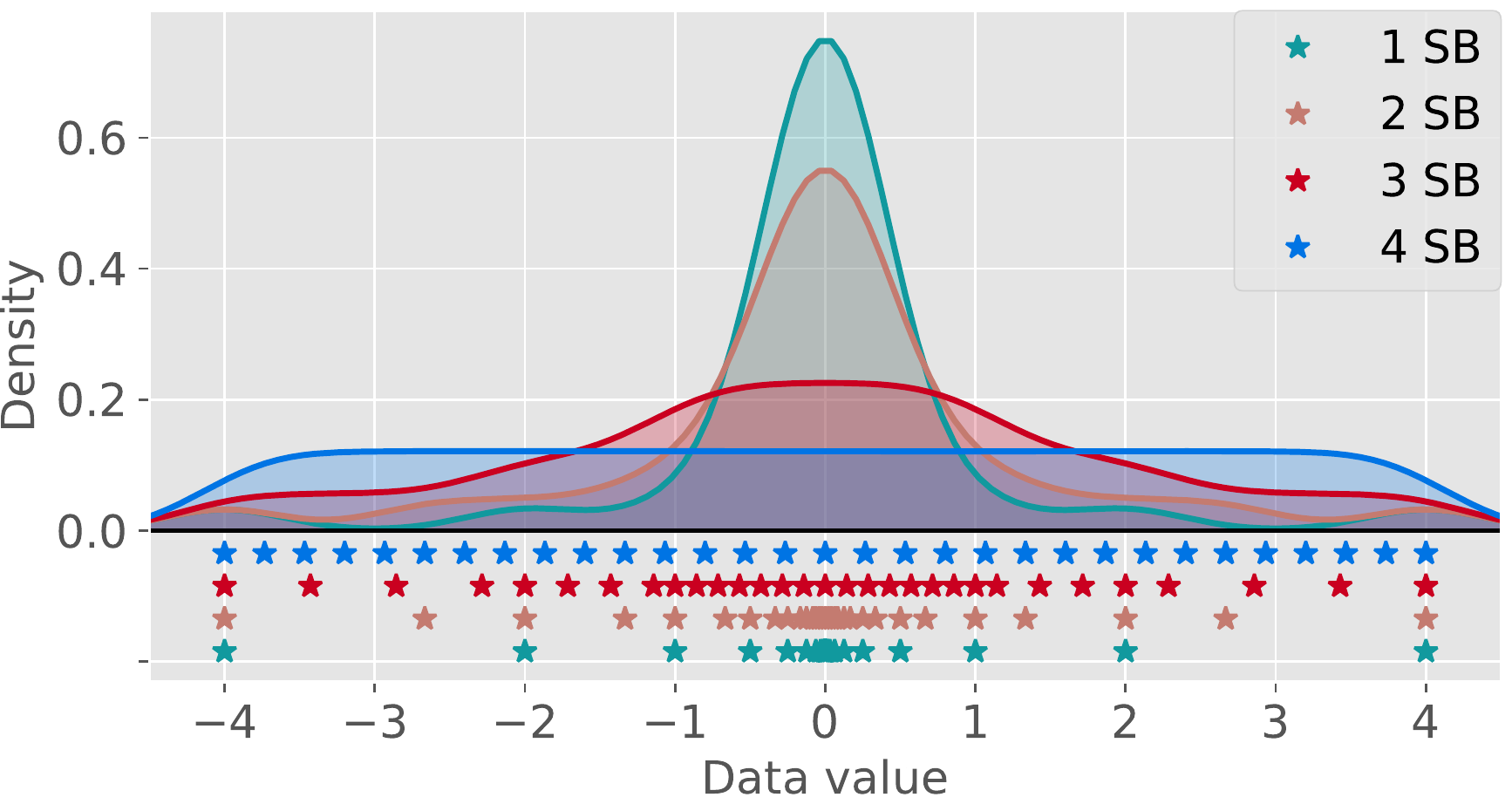}
    }
    \vspace{-5pt}
    \caption{
    Distributions of different quantized values. \highlighted{WD} is the \highlighted{w}eight \highlighted{d}istribution of the first convolutional layer of ResNet18~\cite{he2016resnet} with floating-point data type. \highlighted{QD} denotes \highlighted{q}uantized value \highlighted{d}istribution and \highlighted{QV} indicates \highlighted{q}uantized \highlighted{v}alues. \highlighted{SB} indicates \highlighted{s}ignificant \highlighted{b}its.}
    \label{fig:match_distribution}
\vspace{-15pt}
\end{figure*}
\section{Introduction}
\label{sec:introduction}
\IEEEPARstart{D}{eep} neural networks (DNNs) usually involve hundreds of millions of trained parameters and floating-point operations~\cite{alexnet,he2016resnet}.
Thus, it is challenging for devices with limited hardware resources and constrained power budgets to use DNNs~\cite{PatDNN}.
Quantization has been proven to be an effective method for improving the computing efficiency of DNNs~\cite{rastegari2016xnor,chen2019tdla}. The method maps the full-precision floating-point weights or activation into low-precision representations to reduce the memory footprint of model and resource usage of multiply-accumulate unit~\cite{zhou2016dorefa,DSQ,cheng2019uL2Q}.
For example, the size of a 32-bit floating-point DNN model can be reduced by $32\times$ by quantizing its weights into binaries.

However, low-bitwidth quantization can lead to large error, thus resulting in significant accuracy degradation~\cite{VecQ,PostQ,SYQ}.
The error is the difference of weights/activation before and after quantization which can be quantitatively measured by Euclidean distance~\cite{PostQ}.
Although 8-bit quantization, such as QAT~\cite{QAT} and $\mu$L2Q~\cite{cheng2019uL2Q}, can mitigate the accuracy degradation, use of a high bitwidth incurs comparatively large overheads.
It is still challenging to strive for high accuracy when using a small number of quantizing bits.

Preliminary studies~\cite{LogQ,AddNet,APoT2020} imply that fitting weight distribution of DNNs can improve accuracy.
Fitting distribution refers to adopting a high resolution, i.e., more quantized values, for the densely distributed data range, such as the data range nearing a mean value, while employing a low resolution, i.e., fewer quantized values, for sparsely distributed data range.
The densely distributed data range contains more weights, and adopting high resolution can significantly reduce the quantization errors for quantizing these weights. 
Correspondingly, employing low resolution to the sparsely distributed data range pays almost no impact on the quantization errors, because there are few weights in this range.
Namely, fitting distribution can reduce quantization errors by finding a set of quantized values which distribute close to the weight distribution. 
Deviating weight distribution can be harmful to quantization errors, thereby decreasing model accuracy.
Therefore, a desirable quantization should meet two requirements: the difference between the two distributions of quantized values and weights, and the number of bits used should both be as small as possible.
In other words, to maintain the original features of DNNs, quantized values should fit the distribution of the original values by using fewer bits. 
The closer the two distributions, the higher the quantized DNN accuracy.

Most of the previous studies cannot fit the distribution of the original weights/activation well.
As shown in \figref{fig:match_distribution}, the distribution of original weights is typically bell-shaped~\cite{han2015deep} in the first convolutional layer of ResNet18~\cite{he2016resnet}. 
With five bits for quantization, the distribution of the quantized values of the power of two (PoT) scheme is sharply long-tailed~\cite{LogQ,INQ2017,ENN2017}, and that of the fixed-point method is uniform~\cite{DSQ,QIL,BCGD}, as shown in \figref{fig:dis_of_pot_quantized_values} and \figref{fig:dis_of_linearq_quantized_values}, respectively. 
We can observe that both PoT and the fixed-point deviate from the bell-shaped distribution. 
Although recent studies~\cite{ABC-Net,AddNet,APoT2020} can reduce accuracy degradation by superposing a series of binaries/PoTs, they still encounter limitations in fitting the distributions of weights/activation well. 

We observe that the number of significant bits among the given can impact the distribution of quantized values.
Significant bits in computer arithmetic refer to the bits that contribute to the measurement resolution of the values.
Values represented by different numbers of significant bits and offsets can present different distributions. It implies that we can fit the distributions of weights and activation with elastic significant bits to reduce the accuracy degradation.
In addition, fewer significant bits of operands incur less multiplication complexity, because the multiply operation is implemented by a group of shift-accumulate operations. The group size is proportional to the number of significant bits of operands.
Therefore, the number of significant bits representing weights/activation can directly affect the accuracy and efficiency of the quantized DNN inference.
This motivates us to utilize the significant bits as a key technique to fit the distribution.

\begin{figure}[!t]
    \centering
    \includegraphics[width=0.48\textwidth]{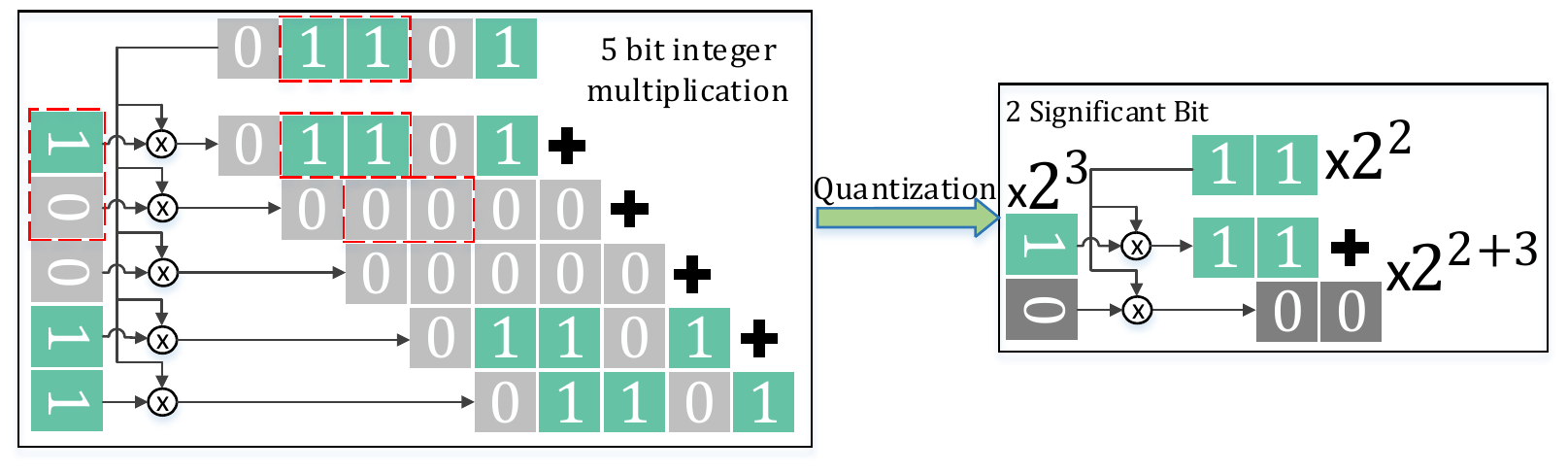}
    \vspace{-10pt}
    \caption{Quantizing values with elastic significant bits can reduce multiplication consumption.}
    \label{fig:computation_with_ESB}
    \vspace{-20pt}
\end{figure}

To demonstrate the advantages of significant bits, we use an example to further clarify our motivation. 
Specifying two significant bits within the five given bits for quantization, we first construct two value sets in which the number of significant bits of values is no more than two: \{0,1\},\{2,3\}.
Then, we enumerate the shifted sets of \{2,3\}, including $2\times\{2,3\}$, $2^2\times\{2,3\}$, $\cdots$, $2^6\times\{2,3\}$, until the number of elements of all sets reaches $2^4$. 
After merging these sets and adding the symmetric negative values, 
we can obtain the set of $\{0,$ $\cdots,$ $\pm 32,$ $\pm 48,$ $\pm 64,$ $\pm 96,$ $\pm 128,$ $\pm 192\}$ with $2^5-1$ elements.
As shown in \figref{fig:dis_of_2sb_quantized_values}, the distribution of the constructed set can fit the weight distribution well, which implies that quantizing DNNs using these quantized values can yield higher accuracy.
In addition, multiplication of the values from the set only requires two shift-accumulate operations, as shown in \figref{fig:computation_with_ESB}, which is the same as the 2-bit quaternary multiplication~\cite{cheng2019uL2Q,VecQ}.
Namely, quantization using two significant bits can achieve high accuracy as the 5-bit fixed-point scheme, but only requires a similar consumption with the multiplication of 2-bit quaternary.

In this study, we propose elastic significant bit (\ours) quantization to achieve high accuracy with few resources by controlling the number of significant bits of quantized values.
\ours~involves three parts: 1) the mathematical formula describing values with a flexible number of significant bits unifies various quantized value set featuring diverse distributions, as shown in \figref{fig:dis_of_esb_quantized_values}; 
2) the distribution difference aligner (DDA) evaluating the matching degree between distributions seeks the desirable quantized value set which fits the original distribution; 
3) \ours~float format coding quantized values in hardware implementation realizes an efficient multiplication computation with low resource consumption.
Our contributions are summarized as follows:
\begin{itemize}
\item To the best of our knowledge, we propose the first unified mathematical expression to describe the quantized values with a flexible number of significant bits, and design a hardware-friendly projection for mapping from full-precision values to these quantized values at runtime.
This formula unifies diverse quantized values, including both PoT and fixed-point forms, and make the values suitable for various distributions, such as the uniform and the bell-shaped ones.
\item We introduce DDA to evaluate the matching degree of distributions before and after quantization.
Based on DDA, the desirable distribution of quantized values, fitting the bell-shaped weights and activation of DNNs, can be found offline without incurring additional computations. 
\item We design \ours~float format to implement the accelerator based on FPGAs, which codes and stores the fraction and exponent of \ours~quantized values separately. This format enables the accelerator to employ low-precision fraction multiplication and shift operation, thus reducing resource consumption and improving energy efficiency.
\item Experimental evaluations on the ImageNet dataset with AlexNet, ResNet18, and MobileNetV2 demonstrated that \ours~achieves higher accuracy than 15 state-of-the-art approaches by \alexnetOimp, \resnetOimp, and \mobileOimp~on average, respectively. 
The \ours~accelerator on FPGAs can achieve a peak performance of 10.95 GOPS/kLUTs without DSPs and improve the energy efficiency by up to \highlighted{65$\times$}, \highlighted{11$\times$} and \highlighted{26$\times$} compared with CPU, GPU, and \sArt~accelerators, respectively.
\end{itemize}

\section{Related works}\label{sec:relatedworks}
DNNs have been widely applied in various areas~\cite{CNN-for-skin-lesions,DL-based-polar-emotion-classification,DL-bigdatama,DRL-for-MRN}, but the high-bitwidth floating-point operations in DNNs affect the inference efficiency seriously.
Early studies on quantization design ultra-low-bitwidth methods to improve the inference efficiency of DNNs, such as binary and ternary methods~\cite{rastegari2016xnor,TSQ2018}.
These studies achieve high computing efficiency with the help of a low bitwidth but failed to guarantee accuracy. Thus, researchers have focused on multi-bit methods to prevent accuracy degradation.

\cusparagraph{Linear quantization.}
The linear quantization quantizes the data into consecutive fixed-point values with uniform intervals. 
Ternarized hardware deep learning accelerator (T-DLA) \cite{chen2019tdla} argues that there are invalid bits in the heads of the binary strings of tensors. It drops both head invalid bits and tail bits in the binary strings of activation. 
Two-step quantization (TSQ) \cite{TSQ2018} uniformly quantizes the activation after zeroing the small values and then quantizes the convolutional kernels into the ternaries. 
Blended coarse gradient descent (BCGD) \cite{BCGD} uniformly quantizes all weights and the activation of DNNs into fixed-point values with shared scaling factors.
Parameterized clipping activation (PACT)~\cite{PACT} parameterizes the quantizing range. Furthermore, it learns the proper range during DNN training and linearly quantizes activation within the learned range. 
Quantization-interval-learning (QIL) \cite{QIL} parameterizes the width of the intervals of linear quantization and learns the quantization strategy by optimizing the DNN loss.
Differentiable Soft Quantization (DSQ) \cite{DSQ} proposes a differentiable linear quantization method to resolve the non-differentiable problem, applying the $tanh$ function to approximate the step function of quantization by iterations.
Hardware-Aware Automated Quantization (HAQ)~\cite{wang2019haq} leverages reinforcement learning (RL) to search the desired bitwidths for different layers and then linearly quantizes the weights and activation of the layers into fixed-point values.

Despite these significant advances in linear quantization, it is difficult to achieve the desired trade-off between efficiency and accuracy owing to the ordinary efficiency of fixed-point multiplication and mismatching nonuniform distributions.

\cusparagraph{Non-linear quantization.}
Non-linear quantization projects full-precision values into low-precision values that have non-uniform intervals. 
Residual-based methods~\cite{he2016resnet,ghasemzadeh2018rebnet} iteratively quantize the residual errors produced by the last quantizing process to binaries with full-precision scaling factors. 
Similar methods, such as learned quantization networks (LQ-Nets) \cite{LQ-Nets}, accurate-binary-convolutional network (ABC-Net) \cite{ABC-Net}, and AutoQ~\cite{AutoQ}, quantize the weights or activation into the sum of multiple groups of binary quantization results. 
The multiply operations of the quantized values of residual-based methods can be implemented as multiple groups of bit operations~\cite{rastegari2016xnor} and their floating-point accumulate operations.
The computing bottleneck of their multiply operations depends on the number of accumulate operations, which is proportional to the number of quantizing bits.
Float-based methods refer to the separate quantization of the exponent and fraction of floating-point values. 
The block floating-point (BFP)~\cite{BFP} and Flexpoint~\cite{Flexpoint} retain the fraction of all elements in a tensor/block and share the exponent to reduce the average number of quantizing bits. 
Posit schemes~\cite{PositArithmetic,DeepPositron} divide the bit string of values into four regions to produce a wider dynamic range of weights: sign, regime, exponent, and fraction. 
AdaptivFloat~\cite{AdaptivFloat} quantizes full-precision floating-point values into a low-precision floating-point format representation by discretizing the exponent and fraction parts into low-bitwidth values, respectively.
Float-based methods usually present unpredictable accuracy owing to a heuristical quantizing strategy and zero value absence in the floating-point format. Few studies have investigated reducing the accuracy degradation of float-based methods.

\cusparagraph{Quantization fitting distribution.}
An efficient way to reduce accuracy degradation by fitting the distributions in multi-bit quantization has been extensively investigated.
Studies on weight sharing, such studies focused on K-means~\cite{hashnet,han2015deep},
take finite full-precision numbers to quantize weights, and store the frequently used weights. Although the shared weights can reduce the storage consumption, it is difficult to reduce the multiplication complexity of full-precision quantized values.
Recent studies~\cite{cheng2019uL2Q,VecQ} attempt to design a better linear quantizer by adjusting the step size of the linear quantizer to match various weight distributions, thus reducing accuracy degradation. 
However, the linear quantizer cannot fit the nonuniform weights well~\cite{APoT2020}, which inevitably degrades the accuracy.
Quantizing the weights into PoT is expected to fit the non-uniform distributions while eliminating bulky digital multipliers~\cite{LogQ}. 
However, PoT quantization cannot fit the bell-shaped weights well in most DNNs~\cite{APoT2020}, as shown in \figref{fig:dis_of_pot_quantized_values}, thus resulting in significant accuracy degradation~\cite{INQ2017,ENN2017}.
AddNet~\cite{AddNet} and additive powers-of-two quantization (APoT)~\cite{APoT2020} exhaustively enumerate the combinations of several computationally efficient coefficients, such as reconfigurable constant coefficient multipliers (RCCMs) and PoTs, to fit the weight distribution while maintaining efficiency.
However, exhaustive enumeration has a high complexity of $O(n!)$. 
In addition, they require expensive look-up operations to map the full-precision weights into their nearest quantized values, that is, the universal coefficient combinations.
These defects can significantly delay model training.
\section{Elastic significant bit quantization}\label{sec:methodology}
In this section, we introduce the \ours~to explain how the quantized values can be controlled at the bit-wise level to fit various distributions. First, we introduce some preliminaries and then expound our \ours~from two aspects: quantized value set and cheap projection operator.
Second, we introduce the DDA to fit the original value distribution and reduce quantization errors.
Finally, We present how \ours~can be exploited to train DNNs.

\subsection{Preliminaries}\label{sec:preliminaries}
\cusparagraph{Notations.} For simplicity, we use weight quantization as an example to describe the quantization process. 
We denote the original weights of a layer in DNNs as $W_f\in \mathbb{R}^d$, and $d$ indicates its dimension.
$W_q$ represents the quantized weights.

\noindent\cusparagraph{Quantized value set and projection operator.} 
A general quantization process projects full precision values into quantized values through \highlighted{the projection operator} $\Pi$, and these quantized values comprise \highlighted{set} $Q$. $\Pi$ impacts the projection efficiency, and $Q$ reflects the distribution law of quantized values. The quantization can be defined as follows:
\begin{equation}
    W_q=\Pi_Q(W_f)
\end{equation}
Here, $\Pi$ projects each original value in $W_f$ into a quantized value in $Q$ and stores it in $W_q$. For \highlighted{linear quantization}, all weights are quantized into fixed-point values as follows:
\begin{equation} 
Q_l(\alpha,b)=\{0,\pm\alpha,\pm2\alpha,\pm3\alpha\cdots,\pm(2^{b-1}-1)\alpha\}
\end{equation}
For \highlighted{PoT}, all weights are quantized into values in the form of PoT as follows:
\begin{equation}
    Q_p(\alpha,b)=\{0,\pm2^0\alpha,\pm2^1\alpha,\pm2^2\alpha,\cdots,\pm2^{2^{b-1}-1}\alpha\}
\end{equation}
Here, $\alpha$ is the scaling factor, and $b$ represents the number of bits for quantization. 
$Q_l(\alpha,b)$ and $Q_p(\alpha,b)$ cannot fit the bell-shaped distribution well, as shown in \figref{fig:dis_of_linearq_quantized_values} and \figref{fig:dis_of_pot_quantized_values}, leading to inevitable accuracy degradation. 
Projection $\Pi$ adopts the nearest-neighbor strategy. It projects each value in $W_f$ into the nearest quantized value in $Q$ measured by the Euclidean distance. 
Because $Q_l(\alpha,b)$ consists of fixed-point values with uniform intervals, linear quantization projection is implemented economically by combining truncating with rounding operations~\cite{zhou2016dorefa,wang2019haq}.
PoT projection can be implemented using a look-up function~\cite{INQ2017,ENN2017,APoT2020}. 
However, the look-up function has a time complexity of $O(n)$.
It involves $d*(2^b-1)$ comparison operations for projecting $W_f$ into $Q_p(\alpha,b)$.
Hence, PoT projection is more expensive than linear quantization projection.
\begin{figure}[!t]
    \centering
    \includegraphics[width=1\columnwidth]{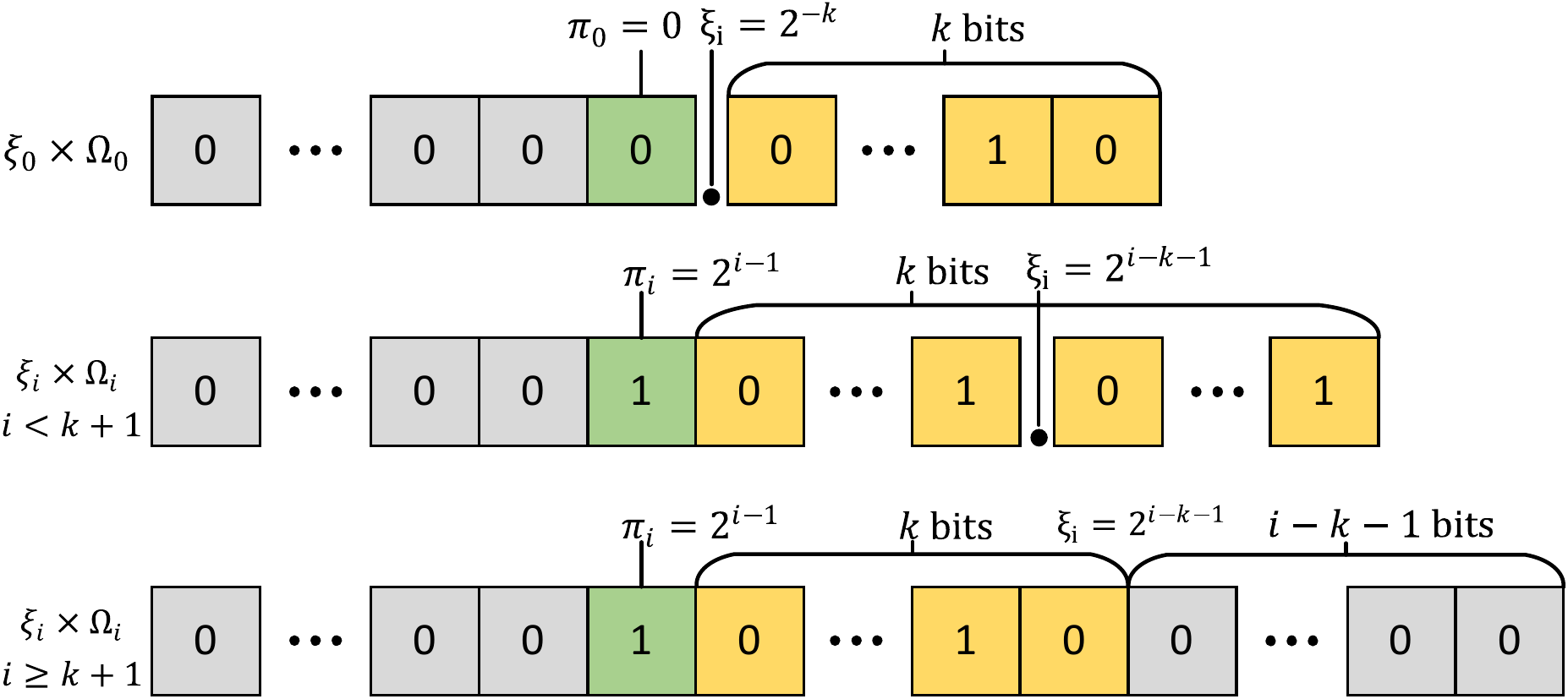}
    \vspace{-15pt}
    \caption{\label{fig:qvalues_less_ksbits}Format of quantized values in \ours.}
    \vspace{-15pt}
\end{figure}

\begin{figure*}[t!]
    \centering
    \includegraphics[width=1\textwidth]{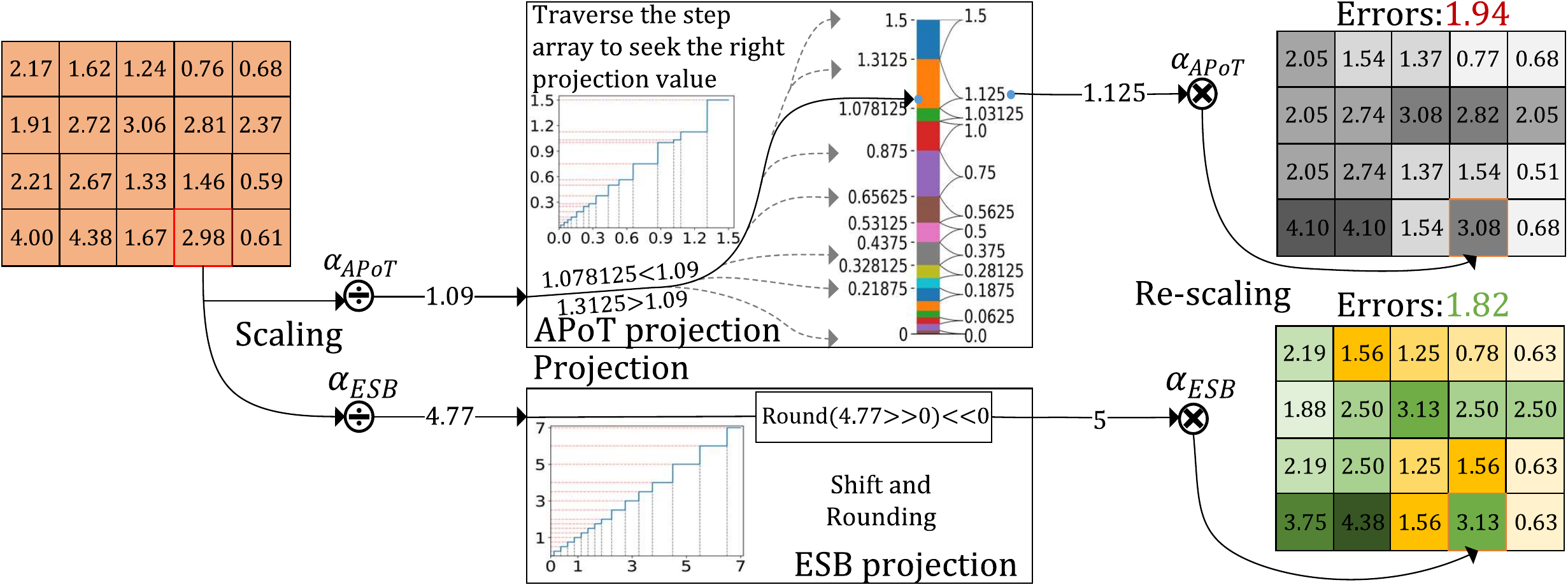}
    \vspace{-15pt}
    \caption{Projection of APoT~\cite{APoT2020} (2 groups with scaling factor $\alpha_{APoT}$) and \ours~($k=2$ and $\alpha_{\ours}$) with five given bits.}
    \label{fig:apot_epot_projection_func}
    \vspace{-15pt}
\end{figure*}
\subsection{Quantized value set}\label{sec:esb_quantized_values}
First, we define the quantized value set and the projection operator for the \ours~as follows.
\begin{equation}
    \label{eq:epot_quantization_expression}
    \begin{split}
    &W_q=\Pi_{Q_e(\alpha,b,k)}(W_f)\\
    &s.t.~Q_e(\alpha,b,k)=\alpha\times(\cup_{i=0}^{N}\xi_i\times\Omega_i),~N=2^{b-k-1}-1
    \end{split}
\end{equation}
Here, $Q_e(\alpha,b,k)$ is the quantized value set, and $\alpha$ is a scaling factor.
$\Omega_i$~($i=0,1,\cdots,N$) is the set of values with the number of significant bits no more than $k+1$, and $\xi_i$ is a scalar. We call $\Omega_i$ the fraction set and $\xi_i$ the shift factor, and we define them as follows.
\begin{equation}
    \begin{split}
    \Omega_i &=\begin{cases}
    \{0,1,2,\cdots,2^k-1\},~i=0\\
    \{2^k,2^k+1,\cdots,2^k+(2^k-1)\},~i>0
    \end{cases}\\
    \xi_i&=\begin{cases}
    2^{-k},~i=0\\
    2^{i-k-1},~i>0
    \end{cases}
    \end{split}
\end{equation}
Here, $k$ is an integer parameter for controlling the number of significant bits of quantized values. $b$ is the number of bits for quantization as mentioned previously. 
Obviously, the number of significant bits of the values in $\Omega_i$ is no more than $k+1$ and $\xi_i$ is a PoT value. 
$Q_e(\alpha,b,k)$ is constructed using the shifted fraction sets $\xi_i\times\Omega_i$.
For ease of notation, we used $\pi=2^k\times\{0,\xi_1,\xi_2,\cdots,\xi_N \}$ to denote the PoT values in $Q_e(\alpha,b,k)$. 
We present the visualized format of the values in $Q_e(\alpha,b,k)$ in \figref{fig:qvalues_less_ksbits}.

We would like to point out that the above mathematical formulas unify all types of distributions, and this is one of our important contributions. In particular, linear quantization can be regarded as a typical case of \ours~quantization when the condition $k=b-2$ is satisfied. Similarly, PoT is equal to the \ours~when $k=0$. Ternary is a special case of \ours~when $k=0$ and $b=2$ are both satisfied.

The multiplication of elements in $Q_e(\alpha,b,k)$ can be converted to the corresponding multiplication of elements in $\Omega_i$ and shift operation (multiply shift factor $\xi_i$).
Therefore, the overhead of matrix multiplications in quantized DNNs with \ours~relies on the number of significant bits of values in $\Omega_i$.
The number is controlled with a configurable parameter $k$ instead of the given bits $b$.
In other words, controlling parameter $k$ in the \ours~can directly impact the computing efficiency of the discretized DNNs.
In addition, we observe that the elastic $k$ can facilitate \ours~to be suitable for various distributions,
ranging from the sharply long-tailed ($k=0$) to the uniform ($k=b-2$) one, including the typical bell-shaped distribution, as shown in \figref{fig:dis_of_esb_quantized_values}. 

Consequently, we can conclude that elastic $k$ has two major effects: fitting the distribution and improving computing efficiency.
By controlling the number of significant bits, the \ours~can achieve a trade-off between accuracy and efficiency.

\subsection{Projection operator}
Next, we shift our attention to projection operator $\Pi$.
We define the full precision value $v$ for the projection operator. $v$ enables a look-up function $P(\cdot)$ based on the nearest neighboring strategy as follows:
\begin{equation}
\small
    \label{eq:thresholing_func}
    \begin{split}
    P(v)&= 
    \begin{cases}
        \pi_i,~v\in[\pi_i-\frac{\xi_{i-1}}{2}, \pi_i+\frac{\xi_i}{2}),~i=0,1,\cdots,N\\
        \tau_i,~v\in[\tau_i-\frac{\xi_i}{2}, \tau_i+\frac{\xi_i}{2}),~i=0,1,\cdots,N
    \end{cases}\\
    &s.t.~v\in [-\mathcal{C},\mathcal{C}]
    \end{split}
\end{equation}
Here, $\tau_i\in\{x: x\in\xi_i\times\Omega_i,x\neq\pi_i\}$, $\mathcal{C}=\max(\xi_N\times\Omega_N)$, and $\max(\cdot)$ returns the maximum element. 
The look-up function in~\eqref{eq:thresholing_func} is convenient for formula expression but expensive in practice, as mentioned in \secref{sec:preliminaries}.

Inspired by the significant bits, we observe that the essence of \ours~projection is to reserve limited significant bits and round the high-precision fraction of values. 
Therefore, we combine the \highlighted{shift} operation with the \highlighted{rounding} operation to implement $P(\cdot)$ and reduce computational overhead.
Specifically, for a value $v\in[-\mathcal{C},\mathcal{C}]$, we seek its most significant 1 bit: $n=\max(\{i: v\ge 2^i\})$.
Let $x(i)$ represent the $i$-th binary valued in $\{0, 1\}$, $v$ can be divided into two parts.
One is $\sum_{i=n-k}^{n}2^{i}\cdot x(i)$, which has $k+1$ significant bits and should be reserved. The other part is $\sum_{i=-\infty}^{n-k-1}2^{i}\cdot x(i)$, which represents the high-precision fraction and should be rounded. 
Therefore, $P(\cdot)$ can be refined as
\begin{equation}
\label{eq:fast_projection}
\begin{split}
P(v)=R(v>>(n-k))<<(n-k)~~~s.t.~v\in [-\mathcal{C},\mathcal{C}]
\end{split}
\end{equation}
Here, the symbols $<<$ and $>>$ represent the left-shift operation and right-shift operation, respectively. $R(\cdot)$ represents \highlighted{rounding} operation. 
All of them are economical computations because they require only one clock cycle computing in modern CPU architectures.
Based on \eqref{eq:fast_projection}, the \ours~projection operator $\Pi$ can be redesigned as follows:
\begin{equation}
    \label{eq:cheap_projection}
    \Pi_{Q_e(\alpha,b,k)}(W_f)=\{\alpha P(v): v\in \lfloor W_f/\alpha \rceil_{-\mathcal{C}}^{\mathcal{C}}\}
\end{equation}
Here, $\lfloor\cdot\rceil_{-\mathcal{C}}^{\mathcal{C}}$ truncates the data in the range of $[-\mathcal{C},\mathcal{C}]$. 
Consequently, this projection operator is beneficial for quantizing activation and weights. 
In particular, the multiplications recognized as the performance bottleneck can also be processed directly, without a projection delay. 

To demonstrate the superiority of our method theoretically, we provide a comparison figure to show the projection process of APoT and \ours.  
As shown in \figref{fig:apot_epot_projection_func}, 
given a vector consists of 20 elements and 5 bits for quantization, APoT (with 2 groups) requires $s=2^5-1$ memory space to pre-store the step-values of the look-up function.
For the input value 2.98, APoT quantizes it into 3.08 through three stages: scaling 2.98 to 1.09 with $\alpha_{APoT}$, costing $s/2$ times on average to look up the right projection value 1.125, and re-scaling 1.125 to 3.08.
Finally, APoT quantizes the original vector to a quantized vector with an error of 1.94, and the time and space complexities of APoT are both $O(s)$. 
In contrast, \ours~($k=2$) does not need to store step-values in advance. \ours~quantizes the value 2.98 through three stages as well: scaling 2.98 to 4.77 with $\alpha_{\ours}$, obtaining the projection value 5 through cheap shift and rounding operations, and re-scaling 5 to 3.13. 
\ours~finally produces a quantized vector with an error of 1.82, and its time and space complexities are only $O(1)$.
Thus, we can conclude that the APoT operator implies $O(s)$ (for storing $s$ steps) for time and space complexity. This leads to high computing overheads compared with our \ours.

\begin{figure*}[!t]
    \centering
    \includegraphics[width=1\textwidth]{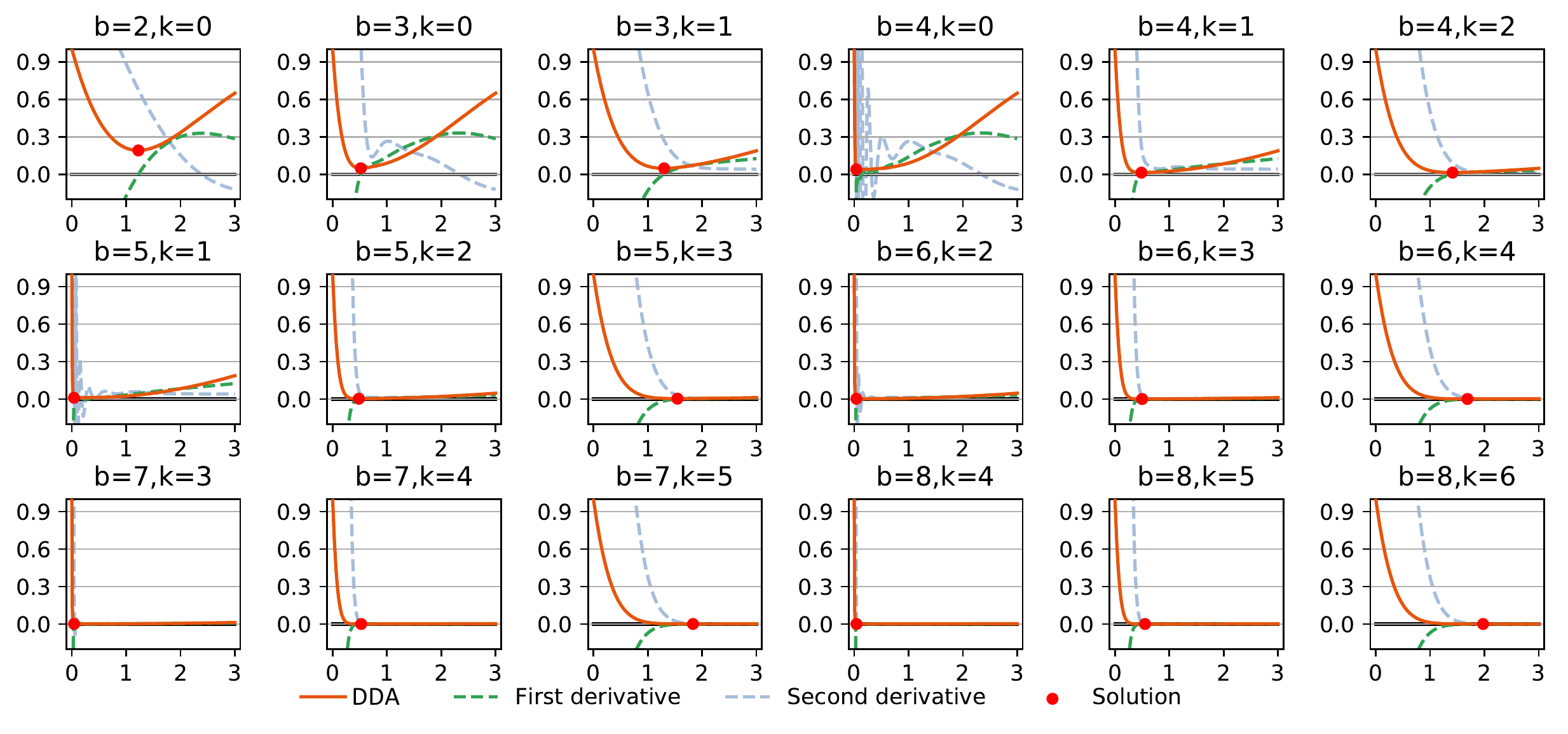}
    \vspace{-25pt}
    \caption{$D(\alpha,b,k)$ with different $\alpha$ and its first and second derivative curves for different values of $b$ and $k$.}
    \label{fig:ddl_extremums}
    \vspace{-15pt}
\end{figure*}
\subsection{Distribution difference aligner}\label{sec:data_normalization}
Assuming that $W_f$ is sampled from a random variable $t$. $p(t;\boldsymbol\theta_{W_f})$ is the probability of $t$, and $\boldsymbol\theta_{W_f}$ is the parameter vector.
For simplicity, we divide this distribution into two parts: the negative ($t<0$) and the non-negative ($t\ge 0$), because the quantized value set is usually symmetric. 
Next, we define the difference $D(\alpha,b,k)$ between the two distributions on the non-negative part as follows:
\begin{equation}
\small
    \label{eq:objective_function_expansion}
    \begin{split}
    &D(\alpha,b,k)=\sum_{q\in Q^*_e(\alpha,b,k)}\int_{t\in \mathcal{S}}(t-q)^2p(t;\theta_{W_f})\mathbf{d}t\\
    &s.t.~\mathcal{S}=
    \begin{cases}
    [0,\alpha\frac{\xi_0}{2}),~q=0\\
    [q-\alpha\frac{\xi_N}{2},\infty),~q=\mathcal{C}\\
    [q-\alpha\frac{\xi_{i-1}}{2},q+\alpha\frac{\xi_{i}}{2}],~q=\alpha\pi_i,~i=0,1,\cdots,N\\
    [q-\alpha\frac{\xi_i}{2},q+\alpha\frac{\xi_i}{2}),~q=\alpha\tau_i,~i=0,1,\cdots,N
    \end{cases}
    \end{split}
\end{equation}
The symbol $^*$ indicates the non-negative part of the set. 
Generally, for a symmetric distribution, we can utilize $D(\alpha,b,k)$ above 
instead of the entire distribution to evaluate distribution difference. 
The challenge of evaluating and optimizing $D(\alpha,b,k)$ is to estimate the probability distribution $p(t;\boldsymbol\theta_{W_f})$ of weights. 
Here, we employ data normalization (DN) and parameterized probability density estimation to address this challenge.

\noindent\cusparagraph{Normalization.} Data normalization (DN) has been widely used in previous works~\cite{APoT2020, ioffe2015batch}, refining data distribution with zero mean and unit variance. Let $\mu$ and $\sigma$ denote the data mean and standard derivation, respectively, weight normalization can be defined as follows:
\begin{equation}
\small
    \begin{split}
    \hat{W}_f&=\frac{W_f-\mu_{W_f}}{\sigma_{W_f}+\epsilon}\\
    &s.t.~\begin{cases}
        \mu_{W_f}=\frac{1}{d}\sum_{j=1}^{d}W_f(j)\\
        \sigma_{W_f}=\sqrt{\frac{1}{d}\sum_{j=1}^{d}(W_f(j)-\mu_{W_f})^2}
    \end{cases}
    \end{split}
\end{equation}
Here, $\epsilon=1e-7$ is a small value to avoid overflow of division.
Estimating the mean and standard deviation of activation results in heavy workloads.
We employ an \highlighted{exponential moving average}~\cite{ioffe2015batch} for parameter estimation of activation on the entire training set in the model training phase. 
For the $i$-th training batch with activation, $A_{fi}$, $\mu_{A_{fi}}$, and $\sigma_{A_{fi}}$ are computed as follows.
\begin{equation}
\small
\begin{cases}
\mu_{A_f}&=(1-\gamma)\mu_{A_f}+\gamma\mu_{A_{fi}}\\
&s.t.~\mu_{A_{fi}}=\frac{1}{d}\sum_{j=1}^{d}A_{fi}(j)\\
\sigma_{A_f}&=(1-\gamma)\sigma_{A_f}+\gamma\sigma_{A_{fi}}\\
&s.t.~\sigma_{A_{fi}}=\sqrt{\frac{1}{d}\sum_{j=1}^{d}(A_{fi}(j)-\mu_{A_{fi}})^2}
\end{cases}
\end{equation}
Here, $\gamma$ is a momentum coefficient and is typically set to 0.9 or 0.99.
$\mu_{A_f}$ and $\sigma_{A_f}$ are the estimated mean and standard deviation of activation on the entire training set, respectively.
The definitions of activation normalization for $A_{fi}$ in the training and test phases are as follows:
\begin{equation}
\small
    \mathrm{Training}:
    ~\hat{A}_{fi}=\frac{A_{fi}-\mu_{A_{fi}}}{\sigma_{A_{fi}}+\epsilon}
    ~~
    \mathrm{Test}:
    ~\hat{A}_{fi}=\frac{A_{fi}-\mu_{A_f}}{\sigma_{A_f}+\epsilon}
\end{equation}

\begin{figure*}[!t]
\centering
    \includegraphics[width=1.0\textwidth]{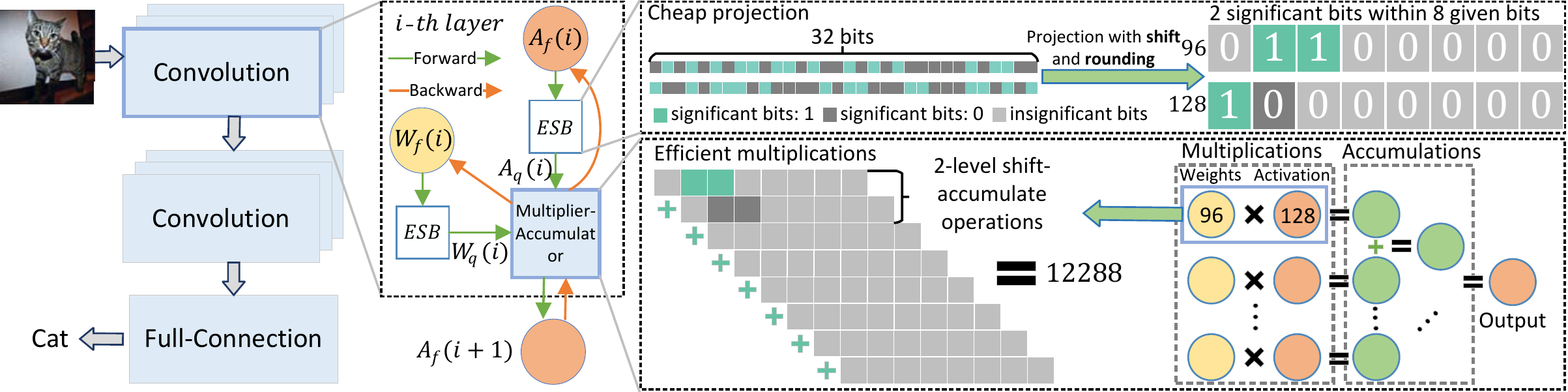}
    \vspace{-18pt}
    \caption{Overall training flow of DNNs with~\ours~quantization.}
    \label{fig:shift_addtion_for_multiplication}
    \vspace{-15pt}
\end{figure*}
\renewcommand{\arraystretch}{1.0}
\begin{algorithm}[!b]
\caption{\label{alg:dnn_training_mlnq}DNN training with \ours.}
\label{alg:neq} 
\begin{algorithmic}[1] 
\REQUIRE Iteration times $T$. Dataset $D$. DNN depth $L$. Learning rate $\eta$.
\ENSURE The trained DNN model.
\FOR{t=1 to T}
    \STATE $A_q(0)=D(t)$
    \FOR{l=1 to L}
        \STATE $\hat{A}_f(l-1)=\text{DN}(A_f(l-1)),~\hat{W}_f(l)=\text{DN}(W_f(l))$
        \STATE $A_q(l-1)=\text{\ours}(\hat{A}_f(l-1)),~W_q(l)=\text{\ours}(\hat{W}_f(l))$
        \STATE $A_f(l)=\text{Forward}(A_q(l-1),W_q(l))$
    \ENDFOR
    \STATE Computing the model loss $Loss$.
    \STATE $G_A(L)=\frac{\partial Loss}{\partial A_f(L)}$
    \FOR{l=L to 1}
        \STATE $G_A(l-1)=G_A(l)
        \cdot\frac{\partial A_f(l)}{\partial A_q(l-1)}
        \cdot\frac{\partial\hat{A}_f(l-1)}{\partial A_f(l-1)}$
        \STATE $G_W(l)=G_A(l)
        \cdot\frac{\partial A_f(l)}{\partial W_q(l)}
        \cdot\frac{\partial\hat{W}_f(l)}{\partial W_f(l)}$
    \ENDFOR
    \FOR{l=1 to L}
        \STATE $W_f(l)=W_f(l)-\eta\cdot G_W(l)$
    \ENDFOR
\ENDFOR
\end{algorithmic}
\end{algorithm}
\noindent\cusparagraph{Resolving extremum.}
Without loss of generality, we assume that all weights and activation of each layer in the DNNs satisfy the normal distribution~\cite{cheng2019uL2Q,TSQ2018}.
Through the above normalization steps, the distribution parameter vector $\boldsymbol\theta_{W_f}$ is $[0,1]^T$. 
Thus, the probability distribution is: $p(t;\boldsymbol{\theta}_{W_f})=\mathcal{N}(t;0,1)$.
After defining $p(t;\boldsymbol{\theta}_{W_f})$, 
all the parts of \eqref{eq:objective_function_expansion} are differentiable, and our optimization can be defined as follows:
\begin{equation}
\label{eq:optimal_alpha_seeking}
    \alpha_{b,k}^*=\arg\underset{\alpha_{b,k}}{\min}(D(\alpha,b,k))
\end{equation}
As shown in \figref{fig:ddl_extremums}, 
we draw the $D(\alpha,b,k)$ curves for $\alpha$ across various conditions.
The horizontal axis represents the value of $\alpha$, and the vertical axis represents $D(\alpha,b,k)$.
When $\alpha$ values are obtained from the given range (0,3], there exist some convex curves such as $D(\alpha, 2,0)$, $D(\alpha, 3,0)$, $D(\alpha, 3,1)$, $D(\alpha, 4,1)$, etc. 
These convex curves imply that we can resolve their extremums through mathematical methods as optimal solutions. 
For non-convex curves, we can still obtain their extremums as feasible solutions which are close to the optimal solutions.  
Therefore, we can directly resolve the extremum of $D(\alpha,b,k)$ to obtain a solution $\alpha_{b,k}^*$ to achieve accurate quantization.
We have computed all desirable solutions $\alpha^*_{b,k}$ for different values of $b$ and $k$. 
The solutions are presented in \tabref{tab:EPoT_under_various_configures}.

\subsection{Training flow}\label{sec:training_flow}
For DNN training, we present the overall training flow as shown in \figref{fig:shift_addtion_for_multiplication}.
Let DNN depth be $L$, and the 32-bit floating-point weights and activation of the $i$-th layer be $W_f(i)$ and $A_f(i)$, respectively.
We embed the \ours~quantization before the multiplier accumulator (MAC) and quantize $W_f(i)$ and $A_f(i)$ into low-precision $W_q(i)$ and $A_q(i)$, respectively.
Then, $W_q(i)$ and $A_q(i)$ are fed into the MAC for acceleration. 
In backward propagation, we employ straight-through-estimation (STE)~\cite{zhou2016dorefa} to propagate the gradients. 
The details are described in \algref{alg:dnn_training_mlnq}.

\begin{figure}[!t]
    \centering
    \includegraphics[width=1\columnwidth]{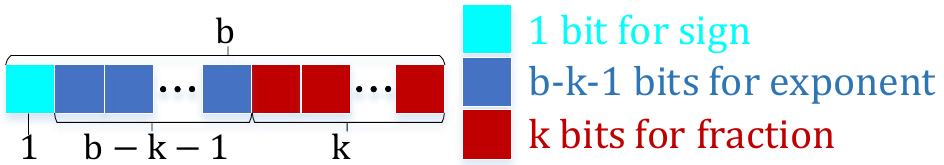}
    \vspace{-17pt}
    \caption{\ours~float format.}
    \label{fig:esb_format_values}
    \vspace{-17pt}
\end{figure}
\section{Hardware implementation}\label{sec:implementation}
The efficient multiplication implementation of ESB quantized values relies on a flexible MAC design, such as low-bitwidth multiplication and shift operation designs. 
However, ESB quantized values cannot be processed efficiently on current general-purpose CPU/GPU platforms, since they do not provide the corresponding instructions for processing these values with flexible bitwidth configurations.

\begin{figure*}
    \centering
    \includegraphics[width=1\textwidth]{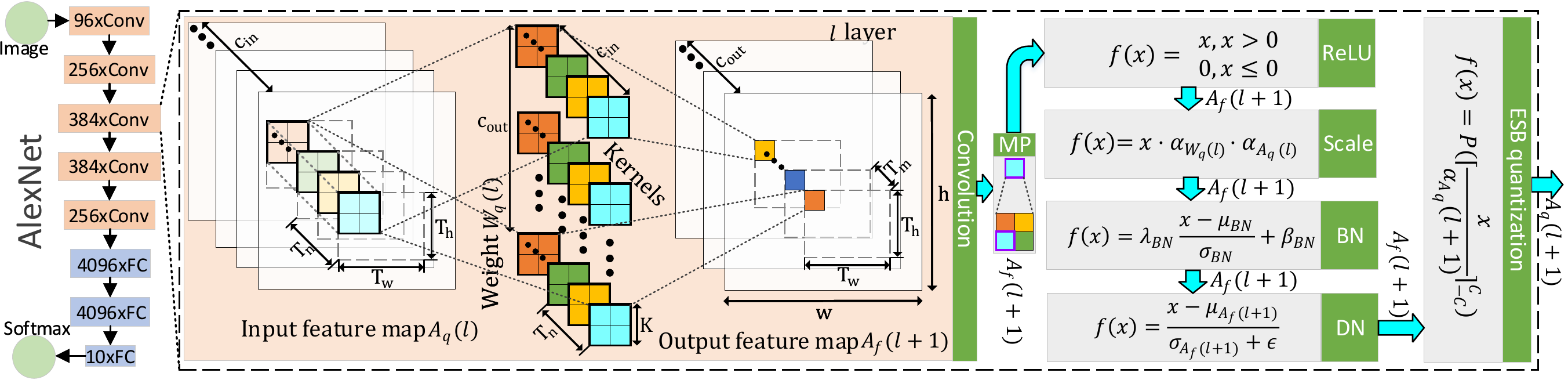}
    \vspace{-20pt}
    \caption{\label{fig:computing_logic}Computing logic of a convolution layer in AlexNet.}
    \vspace{-10pt}
\end{figure*}

\begin{figure*}
    \centering
    \includegraphics[width=1\textwidth]{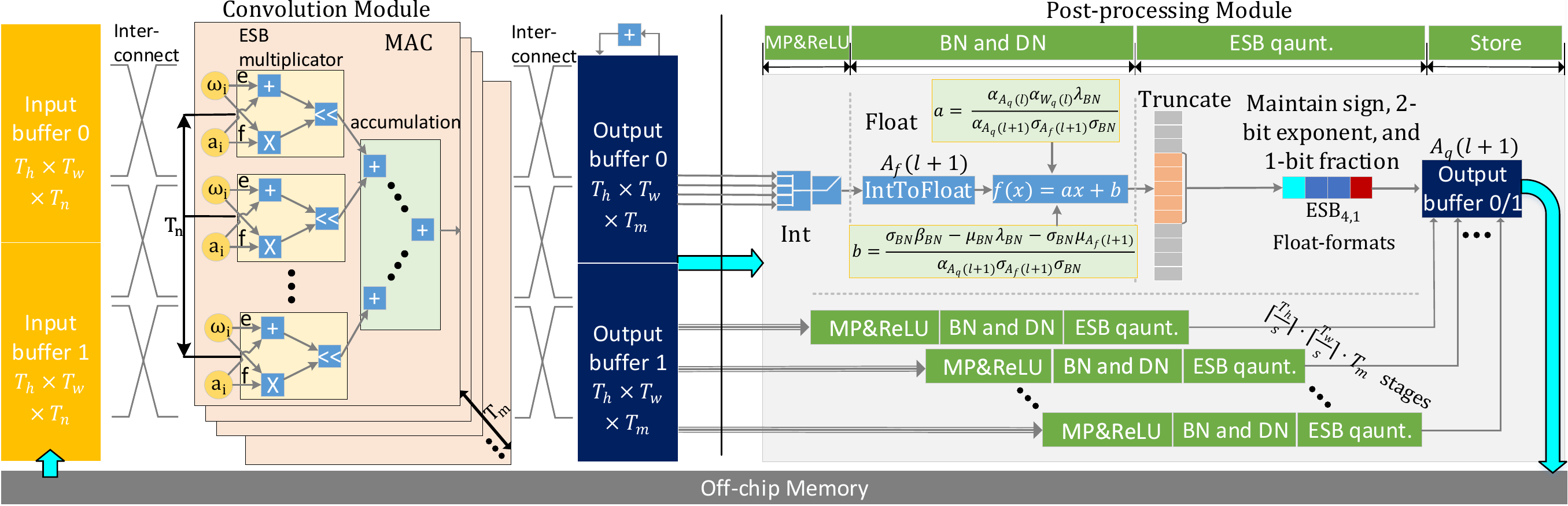}
    \vspace{-18pt}
    \caption{\label{fig:hardware_architecture}Hardware architecture of convolution layer implementation}
    \vspace{-15pt}
\end{figure*}

Consequently, in this section, we introduce the hardware architecture design of the \ours~based on the Field Programmable Gate Array (FPGA) platform. 
We first present the float format for representing \ours~quantized values and explain the multiplication process with \ours\ float format values.
Then, we present the overall hardware architecture design, which includes
1) the general computing logic of DNNs (taking AlexNet as an example) as shown in \figref{fig:computing_logic},
2) hardware architecture as shown in \figref{fig:hardware_architecture}, and
3) timing graph as shown in \figref{fig:timing_hardware_runtime}.
\begin{figure}[!t]
    \centering
    \includegraphics[width=1\columnwidth]{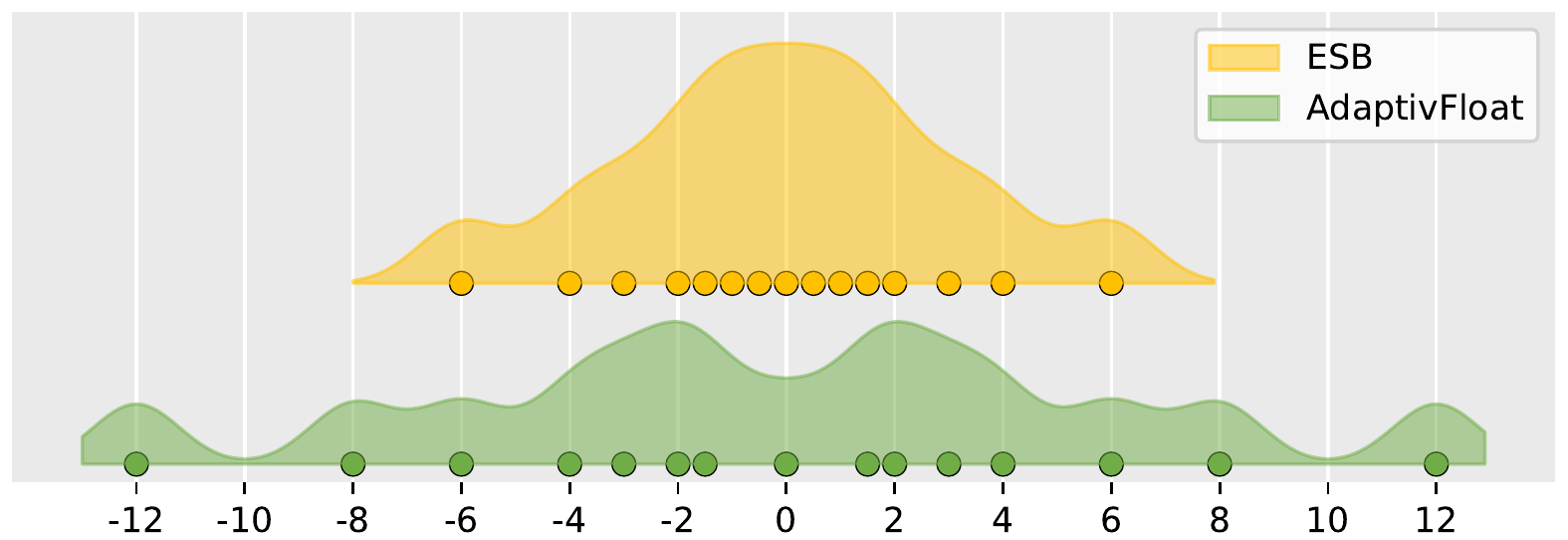}
    \vspace{-15pt}
    \caption{Quantized values of \ours~and AdaptivFloat with the same 1-bit sign, 2-bit exponent and 1-bit fraction.}
    \label{fig:esb_format_dsitribution}
    \vspace{-15pt}
\end{figure}
\subsection{\ours~float format}
The \ours~float format was designed as shown in \figref{fig:esb_format_values}.
It is similar to AdaptivFloat~\cite{AdaptivFloat} (IEEE 754 format), that is, 1 bit for the sign, $k$ bits for the fraction, and $b-k-1$ bits for the exponent. 
Let $e$ be the exponent value, $f$ be the fraction value, and $sb$ be the sign bit (0 or 1); thus, a quantized value $q$ can be represented as follows:
\begin{equation}
\begin{split}
    q=\begin{cases}
    (-1)^{sb}*0.f, &e=2^{b-k-1}-1\\
    (-1)^{sb}*2^e*1.f, &otherwise
    \end{cases}
\end{split}
\end{equation}
Here, we use the representations with a maximum exponent value of $2^{b-k-1}-1$ to represent small numbers around zero.
The difference from AdaptivFloat is that the design of the \ours~float format presents smaller and gradually progressive increasing gaps around zero.
AdaptivFloat cannot represent zero accurately, or it forces the absolute minimum values to zero, resulting in evident larger gaps around zero and losing large information, as shown in \figref{fig:esb_format_dsitribution}.
By contrast, 
\ours\ can present progressively increasing gaps around zero, thus facilitating a smooth distribution around zero and better fitting the weight/activation distributions.
The multiplication of two given values, such as $q_i$ with $sb_i,e_i, f_i$ and $q_j$ with $sb_j,e_j,f_j$, can be derived as follows:
\begin{equation}
\label{eq:esb_multiplication}
    \begin{split}
        q_i\cdot q_j&=(-1)^{sb_i\oplus sb_j}\cdot(f<<e)\\
         &s.t.~
         \begin{cases}
         f&= (\zeta_i+0.f_i)\cdot(\zeta_j+0.f_j)\\
         e&=e_i\zeta_i+e_j\zeta_j\\
         \zeta_i&=\mathbb{I}_{e_i\neq 2^{b-k-1}-1},\zeta_j=\mathbb{I}_{e_j\neq 2^{b-k-1}-1}
        \end{cases}
    \end{split}
\end{equation}
Here, $\zeta_i+0.f_i$ and $\zeta_j+0.f_j$ are the fraction values.
Therefore, the $b$-bit multiplication of the \ours~float format can be implemented by a $k+1$ bit multiply operation at most. $k+1$ is the number of significant bits of the fraction values.

\subsection{Computing logic} 
We use AlexNet as an example to introduce the typical computing logic of a quantized DNN with \ours.
AlexNet has five convolutional layers and three fully connected layers.
A convolutional layer usually consists of four stages: convolution, pooling (max-pooling, MP), activation (ReLU), and batch normalization (BN).
In \ours, the scale stage (Scale), data normalization (DN) (\secref{sec:data_normalization}) stage, and \ours\ quantization (in \eqref{eq:cheap_projection}) stage are attached to quantize the activation values. 
Therefore, the $l$-th convolution layer of AlexNet includes seven stages in our implementation, as shown in \figref{fig:computing_logic}. 

In the convolution stage, we employ the weight $W_q(l)\in Q_e(\alpha_{W_q(l)},b,k)^{K\times K\times c_{in}\times c_{out}}$ to filter the input feature map $A_q(l)\in Q_e(\alpha_{A_q(l)},b,k)^{h\times w\times c_{in}}$ and output $A_f(l+1)\in R^{h\times w\times c_{out}}$. $W_q(l)$ and $A_q(l)$ are the \ours~float-format tensors.
In particular, a convolution operation traverses the input feature map $A_q(l)$ and multiplies accumulating $c_{in}$ channel features and the corresponding $c_{in}$ kernels to produce the output feature map $A_f(l+1)$.
The total number of operations for a convolution computation is up to $h\cdot w\cdot K\cdot K\cdot c_{in}\cdot c_{out}$. 
The following MP operation employs a window of size $p\times p$ to slide on the feature map by the stride step of $s$ and computes the maximum value within the window during sliding. 
If $s$ is not divisible by $h-p$ or $w-p$, the sliding window cannot be aligned with the feature map. 
We captured the intersection window to compute the MP result.
After the MP operation, the size of the feature map $A_f(l+1)$ degrades to $(\lceil\frac{h}{s}\rceil, \lceil\frac{w}{s}\rceil, c_{out})$.
The following five stages only involve injective operations and are defined as shown in \figref{fig:computing_logic}.
These stages can be fused in the model inference.

\subsection{Hardware architecture}
Our accelerator architecture is shown in \figref{fig:hardware_architecture}, which has two modules: convolution and post-processing. 
The convolution module processes the computation-intensive convolution operation, whereas the post-processing module processes all the remaining stages, including MP, ReLU, Scale, BN, DN, and \ours~quantization.

\begin{figure*}
    \centering
    \includegraphics[width=1\textwidth]{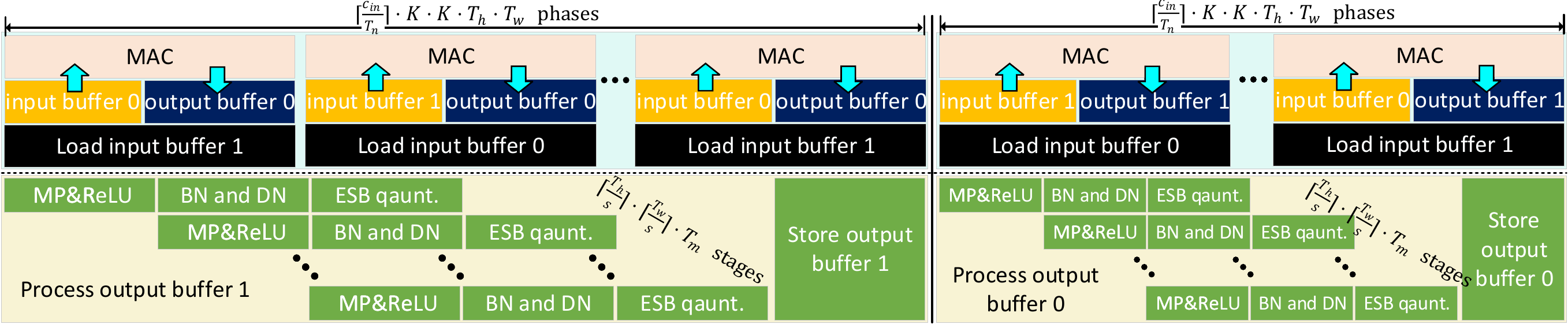}
    \vspace{-20pt}
    \caption{\label{fig:timing_hardware_runtime}Timing graph.}
    \vspace{-18pt}
\end{figure*}
In the convolution module, 
we split the original feature map and output feature map into small tensors with sizes of $(T_h,T_w,T_n)$ and $(T_h, T_w,T_m)$ as shown in \figref{fig:computing_logic}, and split the computing operations into $\lceil\frac{h}{T_h}\rceil\cdot\lceil\frac{w}{T_w}\rceil\cdot\lceil\frac{c_{in}}{T_n}\rceil\cdot\lceil\frac{c_{out}}{T_m}\rceil$ groups of small convolution.
To hide the data transmission time, we employ dual ping-pong input buffers (0/1) of size $(T_h, T_w, T_n)$ and dual ping-pong output buffers (0/1) of size $(T_h, T_w, T_m)$ for a small convolution computation, as shown in \figref{fig:hardware_architecture}.
Then, we design $T_m\cdot T_n$ MAC units, which can execute $2\cdot T_m\cdot T_n$ multiply-accumulating operations in parallel. 
Therefore, we can reduce the computing time of the convolution stage by $T_m\cdot T_n$ times and accelerate the convolution inference.

In the post-processing module, 
we compress the six stages into three steps by fusing some stages into one, as shown in \figref{fig:hardware_architecture}.
The "MP\&ReLU" step fuses MP and ReLU operations. 
MP takes a sliding window with a pooling size of $p\times p$ and a stride step of $s$ to process the small output tensor of size $(T_h, T_w, T_m)$. 
If the sliding window of the MP cannot be aligned with the tensor, we cache their intersection window and delay the pooling operation until the next computing trip.
The following ReLU eliminates negative MP outputs.
Then, we fuse the Scale, BN, DN stages, and the division part of \ours~quantization into a single step of “BN and DN” as follows:
\begin{equation}
\begin{split}
    &\text{DN(BN(Scale(}x\text{)))}/\alpha_{A_q}(l+1)=ax+b\\
    &s.t.~\begin{cases}
    a&=\frac{\alpha_{W_q(l)}\cdot\alpha_{A_q(l)}\cdot\lambda_{BN}}{\alpha_{A_q(l+1)}\cdot\sigma_{A_f(l+1)}\cdot\sigma_{BN}}\\
    b&=\frac{\sigma_{BN}\cdot\beta_{BN}-\mu_{BN}\cdot\lambda_{BN}-\sigma_{BN}\cdot\mu_{A_f(l+1)}}{\alpha_{A_q(l+1)}\cdot\sigma_{A_f(l+1)}\cdot\sigma_{BN}},
    \end{cases}
\end{split}
\end{equation}
Here, $\alpha_{A_q(l)},\alpha_{W_q(l)}$ are the scaling factors of $A_q(l)$ and $W_q(l)$ of the $l$-th layer, respectively. $\alpha_{A_q(l+1)}$ is the scaling factor for $A_q(l+1)$, which is the input of the $(l+1)$-th layer. 
$\mu_{BN}$, $\sigma_{BN}$, $\lambda_{BN}$, and $\beta_{BN}$ are the parameters of BN, $\mu_{A_f(l+1)}$ and $\sigma_{A_f(l+1)}$ are the parameters of DN.
Since all of these parameters are constant in the model inference, we compute $a$ and $b$ offline to reduce the computing overhead. 
The "\ours~quantization" step, implemented with the truncation and projection as shown in \eqref{eq:epot_quantization_expression}, quantizes elements of $A_f(l+1)$ into the \ours~float format and then stores them into $A_q(l+1)$.
We design a pipeline to conduct the three steps.
$A_q(l+1)$ is written to output buffer 0 or 1 and finally stored in the off-chip memory.

\subsection{Timing graph} 
Using the \ours~pipeline design, we can compute the convolution module and post-processing module in parallel to fully utilize hardware resources and accelerate DNN inference.
The timing graph of the accelerator is shown in \figref{fig:timing_hardware_runtime}. 
In the convolution module, we utilize dual ping-pong buffers for input prefetching to hide the data transmission time when executing MAC in the convolution module. 
The dual output buffers facilitate the execution of two modules in parallel. 
When the convolution module writes data into the output buffer 0, the post-processing module deals with another output buffer 1, and vice versa.
\section{Evaluation}\label{sec:experiments}
To validate the unique features of \ours~and demonstrate its accuracy and efficiency benefits, we conducted extensive experiments on representative image classification tasks and implemented \ours~quantization on FPGAs. 

\subsection{Experimental setting}
In the experiments, all the weights and activation of all layers in DNNs, including the first and last layers, are quantized by \ours~with the same settings.
We use Keras v2.2.4 and TensorFlow v1.13.1 as the DNN training frameworks.
The layers in Keras are reconstructed as described in \secref{sec:training_flow} to perform \ours~quantization in the image classification task.

\cusparagraph{Datasets and models.} 
Five DNN models on three data sets are used in our experiments.
We utilize LeNet5 with 32C5-BN-MP2-64C5-BN-MP2-512FC-10Softmax
on MNIST and VGG-like with 64C3-BN-64C3-BN-MP2-128C3-BN-128C3-BN-MP2-256C3-BN-256C3-BN-MP2-1024FC-10-Softmax on Cifar10.
Here, \highlighted{C} is convolution, \highlighted{BN} is batch normalization, \highlighted{MP} indicates maximum pooling, and \highlighted{FC} denotes full connection.
Data augmentation in DSN~\cite{lee2015deeply} 
is applied for VGG-like training.
We also conduct experiments over two conventional networks, \highlighted{AlexNet}~\cite{alexnet} and \highlighted{ResNet18}~\cite{he2016resnet}, and a light-weight network \highlighted{MobileNetV2}~\cite{sandler2018mobilenetv2} on ImageNet~\cite{deng2009imagenet}.
We use the BN~\cite{ioffe2015batch} layer to replace the original local response normalization (LRN) layer in AlexNet for a fair comparison with ENN~\cite{ENN2017}, LQ-Nets~\cite{LQ-Nets}, and QIL~\cite{QIL}.
We adopt the open data augmentations for ImageNet as in TensorFlow: resize images into a size of $224\times 224$ following a random horizontal flip. Then, we subtract the mean of each channel for AlexNet and ResNet18, and scale pixels into the range of [-1,1] for MobileNetV2.

\renewcommand{\arraystretch}{1.0}
\begin{table*}[!t]
\centering
\caption{Results of \ours~across various configurations.}
\label{tab:EPoT_under_various_configures}
\vspace{-7.5pt}
\setlength\tabcolsep{11pt} 
\begin{tabular}{c|c|l|l|c|c|c|c|c|c}\Xhline{1.5pt}
\multicolumn{1}{c|}{\multirow{3}{*}{{W/A}}} & 
\multicolumn{1}{c|}{\multirow{3}{*}{\bm{$k$}}}&
\multicolumn{1}{c|}{\multirow{3}{*}{{Name}}} & 
\multirow{3}{*}{{Alias}} & \multicolumn{1}{c|}{\multirow{3}{*}{{DDA}}} & 
\multicolumn{1}{c|}{\multirow{3}{*}{\bm{$\alpha$}$^*$}}& 
\multicolumn{4}{c}{{Datasets \& Models}} \\\cline{7-10}
 & & & &  &  &\multicolumn{1}{c|}{{Mnist}} & \multicolumn{1}{c|}{{Cifar10}} & \multicolumn{2}{c}{{ImageNet(Top1/Top5)}} \\\cline{7-10}
 & & & &  &  &\multicolumn{1}{c|}{{LeNet5}} & \multicolumn{1}{c|}{{VGG-like}} & \multicolumn{1}{c|}{{AlexNet}}&
 \multicolumn{1}{c}{{ResNet18}}\\\Xhline{1.5pt}
32/32 & - & Referenced & - &  0 & - & 99.40& 93.49&60.01/81.90$\dagger$ & 69.60/89.24$\dagger$ \\\Xhline{0.5pt}
2/2 & 0 & \ours(2,0) & Ternary& \highlighted{0.1902} & 1.2240 & 99.40&90.55 &58.47/80.58 & 67.29/87.23\\\Xhline{0.5pt}
3/3 & 0&\ours(3,0)& PoT& 0.0476 & 0.5181 & 99.45 & 92.55 &60.73/82.53 &69.29/88.79 \\
3/3 & 1 & \ours(3,1)& Fixed& \highlighted{0.0469} & 1.3015 & \highlighted{99.61}&92.76 & 61.52/82.96 & 69.82/89.10 \\\Xhline{0.5pt}
4/4 & 0 &\ours(4,0)& PoT& 0.0384 & 0.0381& 99.61&92.54 &61.02/82.38 & 69.82/89.04 \\
4/4 &1&\ours(4,1)& ESB& \highlighted{0.0127} & 0.4871 & 99.53 &93.41  &62.19/83.37 &70.41/89.43\\
4/4 &2&\ours(4,2)& Fixed& 0.0129 & 1.4136 & 99.52 &93.30 &61.45/82.64 & 70.36/89.29 \\\Xhline{0.5pt}
5/5 &1 &\ours(5,1)& ESB& 0.0106 & 0.0391 &  99.55&93.44 & 61.68/83.08 &70.41/89.57\\
5/5 &2&\ours(5,2)& ESB& \highlighted{0.0033} & 0.4828& 99.53  &93.43 & 62.42/83.52&71.06/89.48 \\
5/5 &3 &\ours(5,3)& Fixed& 0.0037 & 1.5460& 99.54&93.26 &61.72/82.93 &70.60/89.53 \\\Xhline{0.5pt}
6/6 &2 &\ours(6,2)& ESB& 0.0028 & 0.0406& 99.52 &\highlighted{93.53} &62.07/83.31 &70.30/89.58 \\
6/6 &3 &\ours(6,3)& ESB& \highlighted{0.0008} &0.4997& 99.53 &93.49 &62.39/83.58&70.99/89.55\\
6/6 &4&\ours(6,4)& Fixed& 0.0011 & 1.6878& 99.52 &93.40 & 61.91/83.23&70.30/89.37 \\\Xhline{0.5pt}
7/7 &3 &\ours(7,3)& ESB& 0.0007 & 0.0409& 99.52 &93.29 & 61.99/83.32 &\highlighted{71.06/89.55} \\
7/7 &4&\ours(7,4)& ESB& \highlighted{0.0002} & 0.5247&  99.54&93.29 &62.26/83.25 &70.98/89.58\\
7/7 &5&\ours(7,5)& Fixed& 0.0003 & 1.8324& 99.53 &93.32 & 61.84/83.20 &70.91/89.62 \\\Xhline{0.5pt}
8/8 &4&\ours(8,4)& ESB& 0.0002 & 0.0412& 99.52 &93.47 &62.22/83.40 &71.03/89.78 \\
8/8 & 5&\ours(8,5)& ESB& \highlighted{0.0001} & 0.5527& 99.53 & 93.51&\highlighted{62.94/84.15} &70.96/89.74\\
8/8 &6&\ours(8,6)& Fixed& 0.0001 & 1.9757& 99.55 &93.36 &61.63/82.96 &70.96/89.70\\\Xhline{1.5pt}
\end{tabular}\flushleft\vspace{-5pt}
\footnotemark[1]{\highlighted{W/A} denotes the given bits for weight quantization and activation quantization, respectively. \highlighted{Name} is abbreviated.
\highlighted{Alias} refers to special cases of ESB; Fixed denotes the linear quantization.}
\highlighted{DDA} is the distribution difference estimation, and the bold results represent the minimal DDA under the corresponding given bits and significant bits.
\highlighted{$\alpha^*$} indicates a desirable solution.
The configurations of ESB($b,k$) $b-k>4$ are not recommended because they are similar to configurations of ESB($b-1,k$) in terms of accuracy but consume more bits.
$\dagger$ The results refer to \cite{VecQ}.
\vspace{-5pt}
\end{table*}
\renewcommand{\arraystretch}{1.0}
\begin{table*}[!t]
\centering
\caption{\label{tab:comparison_methods_config}
Detailed settings of quantization methods collected from the literature.}
\vspace{-7.5pt}
\setlength\tabcolsep{11pt} 
\begin{tabular}{l|l|l|ccc|ccc|cc}\Xhline{1.5pt}
\multicolumn{2}{c|}{\multirow{2}{*}{{Methods}}} &
\multirow{2}{*}{{Format}} &
\multicolumn{3}{c|}{{Weights}} & \multicolumn{3}{c|}{{Activation}} & \multicolumn{1}{c}{\multirow{2}{*}{{FConv}}} & \multicolumn{1}{c}{\multirow{2}{*}{{LFC}}} \\\cline{4-9}
\multicolumn{2}{c|}{} & & \multicolumn{1}{c}{{Bits}} & \multicolumn{1}{c}{{SFB}} & \multicolumn{1}{c|}{{SFN}} & \multicolumn{1}{c}{{Bits}} & \multicolumn{1}{c}{{SFB}} & \multicolumn{1}{c|}{{SFN}} & \multicolumn{1}{c}{} & \multicolumn{1}{c}{} \\\Xhline{1.5pt}
\multirow{9}{*}{\rotatebox{90}{\highlighted{Linear (Fixed-point)}}}
& $\mu$L2Q~\cite{cheng2019uL2Q} &Fixed& 2,4,8 & 32 & 1 & 32 & -& -& Y  & 8 \\
& VecQ~\cite{VecQ} & Fixed&2,4,8 & 32 & 1 & 8 & 32 & 1  & Y  & Y \\
& TSQ~\cite{TSQ2018} & Fixed&2 & 32 & $c~\dagger$ & 2 & 32 & 2  & N  & N \\
& BCGD~\cite{BCGD} & Fixed&4 & 32 & 1 & 4 & 32 & 1  & Y  & Y \\
& HAQ~\cite{wang2019haq} & Fixed&f2,f3,f4$~\S$ & 32 & 1 & 32,f4 & -/32 & -/1  & 8  & Y \\
& DSQ~\cite{DSQ} &Fixed& 2,3,4 & 32 & 1 & 2,3,4 & 32 & 1  & N  & N \\
& PACT~\cite{PACT} &Fixed& 2,3,4,5 & 32 & 1 & 2,3,4,5 & 32 & 1  & N  & N \\
& Dorefa-Net~\cite{zhou2016dorefa} &Fixed& 2,3,4,5 & 32 & 1 & 2,3,4,5 & 32 & 1  & N  & N \\
& QIL~\cite{QIL} & Fixed&2,3,4,5 & 32 & 1 & 2,3,4,5 & 32 & 1  & N  & N \\
\Xhline{1.5pt}
\multirow{8}{*}{\rotatebox{90}{\highlighted{Nonlinear}}} 
& INQ~\cite{INQ2017} &PoT& 2,3,4,5 & 32 & 1 & 32 &-&-&-&-\\
& ENN~\cite{ENN2017} &PoT& 2,3 & 32 & 1 & 32 & - & -  &  - & - \\
& ABC-Net~\cite{ABC-Net} & -&3,5 & 32 & \{3,5\}*$c$ & 3,5 & 32 & 3,5  & -  & - \\
& LQ-Nets~\cite{LQ-Nets}&-& 2,3,4 & 32 & \{2,3,4\}*$c$ & 2,3,4 & 32 & 2,3,4  & N  & N \\
& AutoQ~\cite{AutoQ}& -&f2,f3,f4 & 32 & \{f2,f3,f4\}*$c$ & f2,f3,f4 & 32 & f2,f3,f4 & - & - \\
& APoT~\cite{APoT2020} & -&2,3,5 & 32 & 1 & 2,3,5 & 32 & 1  & N  & N
\\\cline{2-11}
& \highlighted{\ours} &ESB& {2-8} & {32} & {1} & {2-8} & {32} & {1} & \highlighted{Y} & \highlighted{Y}\\
& \highlighted{\ours*} &ESB& {2-5} & {32} & {1} & {2-5} & {32} & {1} & \highlighted{N} & \highlighted{N}\\\Xhline{1.5pt}
\end{tabular}\\\flushleft\vspace{-5pt}
\footnotemark[1]{\highlighted{SFB} is scaling-factor bitwidth, and \highlighted{SFN} is the number of the scaling factors. \highlighted{FConv} represents the first convolutional layer, \highlighted{LFC} represents the last fully connected layer, and Y or N indicates whether quantization was done.
Fixed refers to the fixed-point format.}
$\dagger$~{$c$ denotes the number of convolutional kernels.}
$\S$~{\highlighted{f$x$} denotes there are $x$ average bits for the mixed-precision schemes.}
\vspace{-15pt}
\end{table*}
\noindent\cusparagraph{Model training.} We take the open pre-trained models from~\cite{VecQ} to initialize the quantized models and then fine-tune them to converge. 
For experiments with fewer bits than 4, we train the quantized models for a total number of 60 epochs with an initial learning rate (LR) of 1.0. LR is decayed to 0.1, 0.01, and 0.001 at 50, 55, and 57 epochs, respectively.
For experiments with given bits of more than 4, we train the quantized models for a total number of 10 epochs with an initial LR of 0.1, and decay the LR to 0.01 and 0.001 at 7 and 9 epochs, respectively.

\cusparagraph{Hardware platform.} 
Model training and evaluation are based on a \highlighted{Super Micro 4028GR-TR} server equipped with \highlighted{two 12-core Intel CPUs (Xeon E5-2678 v3)}, an \highlighted{Nvidia RTX 2080ti GPU}, and \highlighted{64GB memory}.
The accelerator system of the \ours~is implemented on the Xilinx ZCU102 FPGA platform, which is equipped with a dual-core ARM Cortex-A53 and based on Xilinx's 16nm FinFET+ programmable logic fabric with 274.08K LUTs, 548.16K FFs, 912 BRAM blocks with 32.1Mb, and 2520 DSPs. Vivado 2020.1 is used for system verification and \ours~simulation.

\subsection{\ours~accuracy across various bits}
We first focus on the investigation of the inference accuracy achieved by \ours\ across various bits. The notation with the value pair for \ours, such as \highlighted{\ours($b$,$k$)}, indicates that there are $b$ given bits and $k$ significant bits among $b$ bits in the \ours~scheme. 
As mentioned in \secref{sec:esb_quantized_values}, linear quantization ($k=b-2$), PoT ($k=0$), and Ternary ($k=0$ and $b=2$) are special cases of \ours. A series of experimental results are shown in \tabref{tab:EPoT_under_various_configures} to verify the effect of elastic significant bits on the accuracy.

\ours~can retain high accuracy across all datasets and DNN models. 
For example, when pair $(b,k)$ has the values of (3,1), (6,2), (8,5), and (7,3), the accuracy results of \ours~can reach $\highlighted{99.61\%}$, $\highlighted{93.53\%}$, $\highlighted{62.94\%}$(Top1), and $\highlighted{71.06\%}$(Top1), for LeNet5, VGG-like, AlexNet, and ResNet18, respectively, which outperform the referenced pre-trained models by $\highlighted{0.21\%}$, $\highlighted{0.04\%}$, $\highlighted{2.93\%}$, and $\highlighted{1.46\%}$, respectively.
When the conditions $b-k=3~(b>3)$ are satisfied, \ours~has the minimal DDAs while maintaining relatively higher accuracy, such as~\ours(4,1) and~\ours(5,2). 
In particular, under three given bits, the results of \ours~outperform referenced pre-trained AlexNet and ResNet18 by $\highlighted{1.51\%}$ and $\highlighted{0.22\%}$, respectively, in terms of accuracy.
Our proposed DDA calculates the distribution of quantized values of \ours~to fit the bell-shaped distribution well, thus maintaining the features of the original values and reducing the accuracy degradation. 
In other words, the minimal DDA result corresponds to the best quantizing strategy of the \ours~for different given bits.

\begin{figure*}[!t]
    \centering
    \subfloat[\ours~vs. Linear-Q on AlexNet]{
        \includegraphics[width=0.31\textwidth]{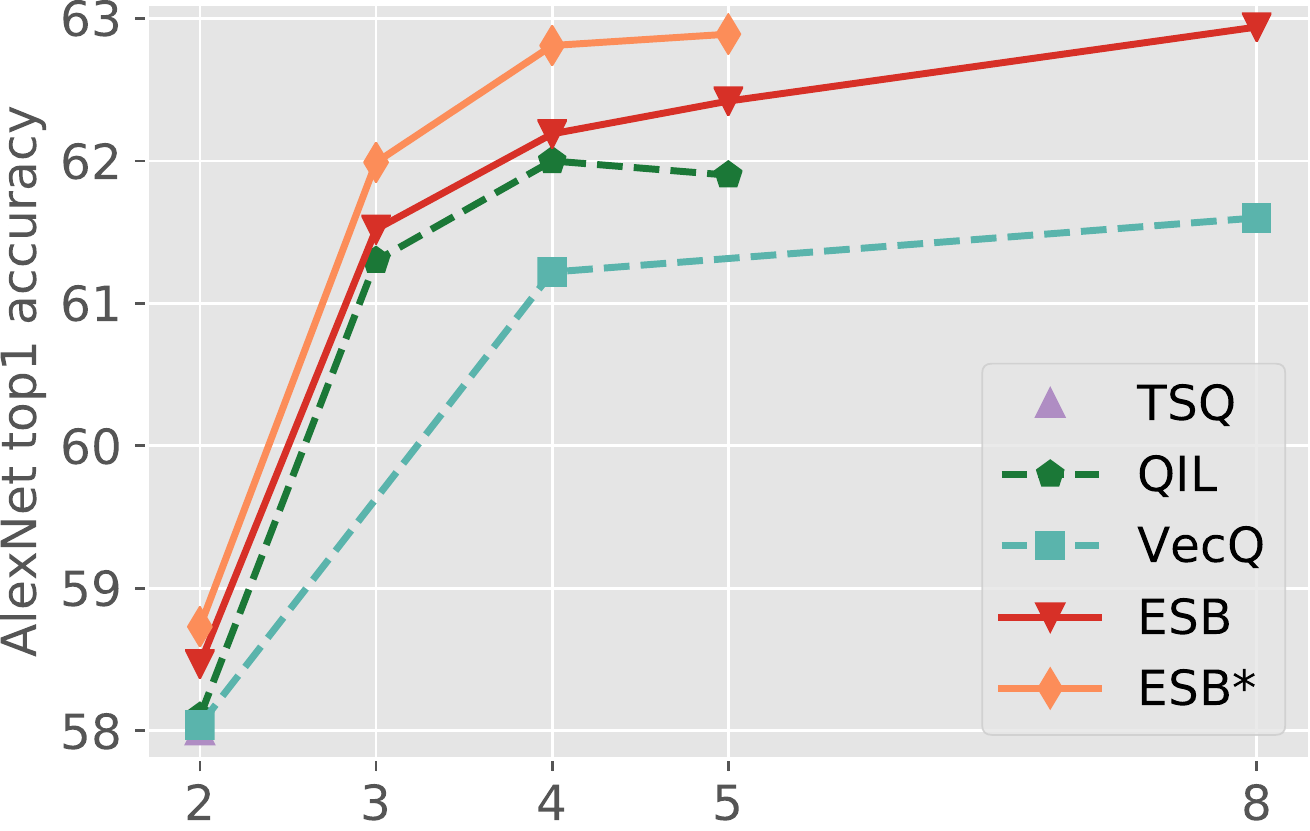}
    }
    \subfloat[\ours~vs. Nonlinear-Q on AlexNet]{
        \includegraphics[width=0.31\textwidth]{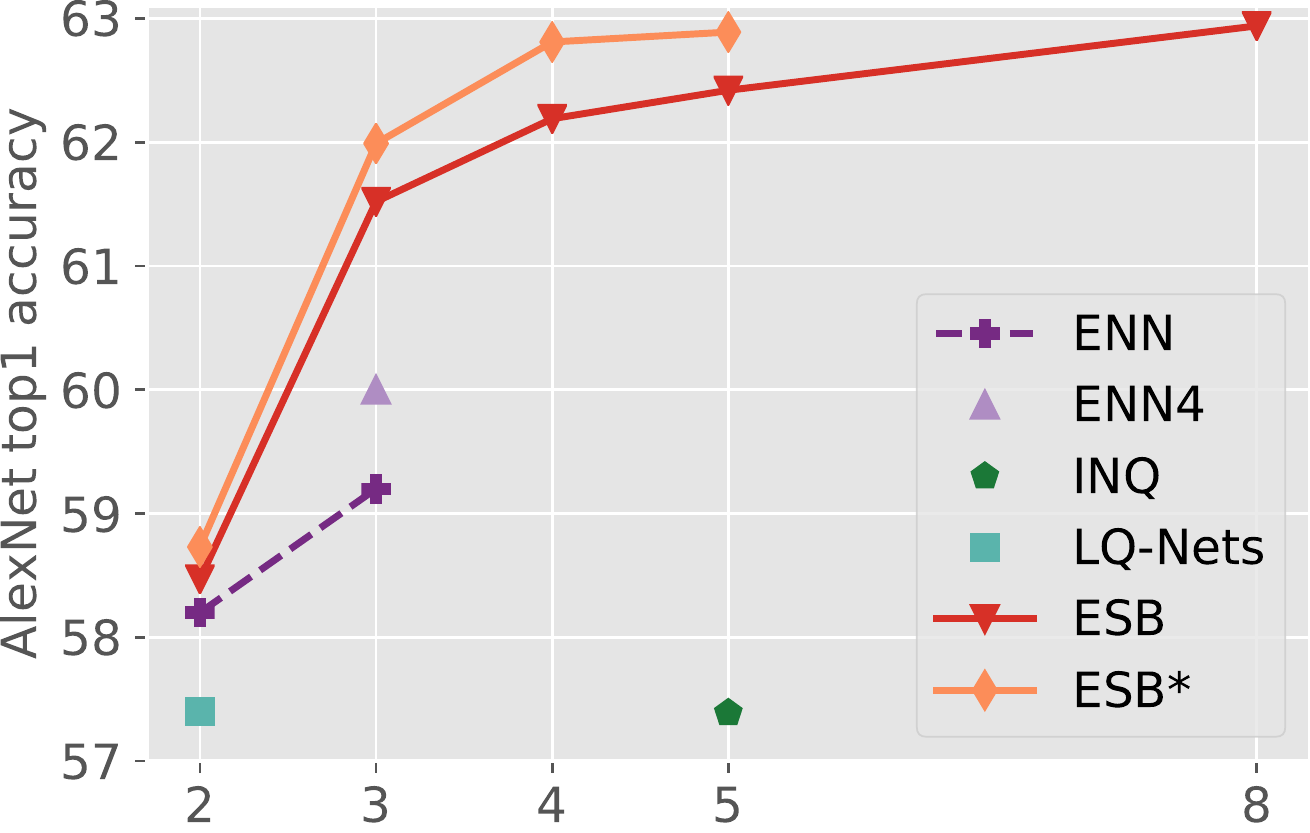}
    }
    \subfloat[\ours~vs. Linear-Q on ResNet18]{
        \includegraphics[width=0.31\textwidth]{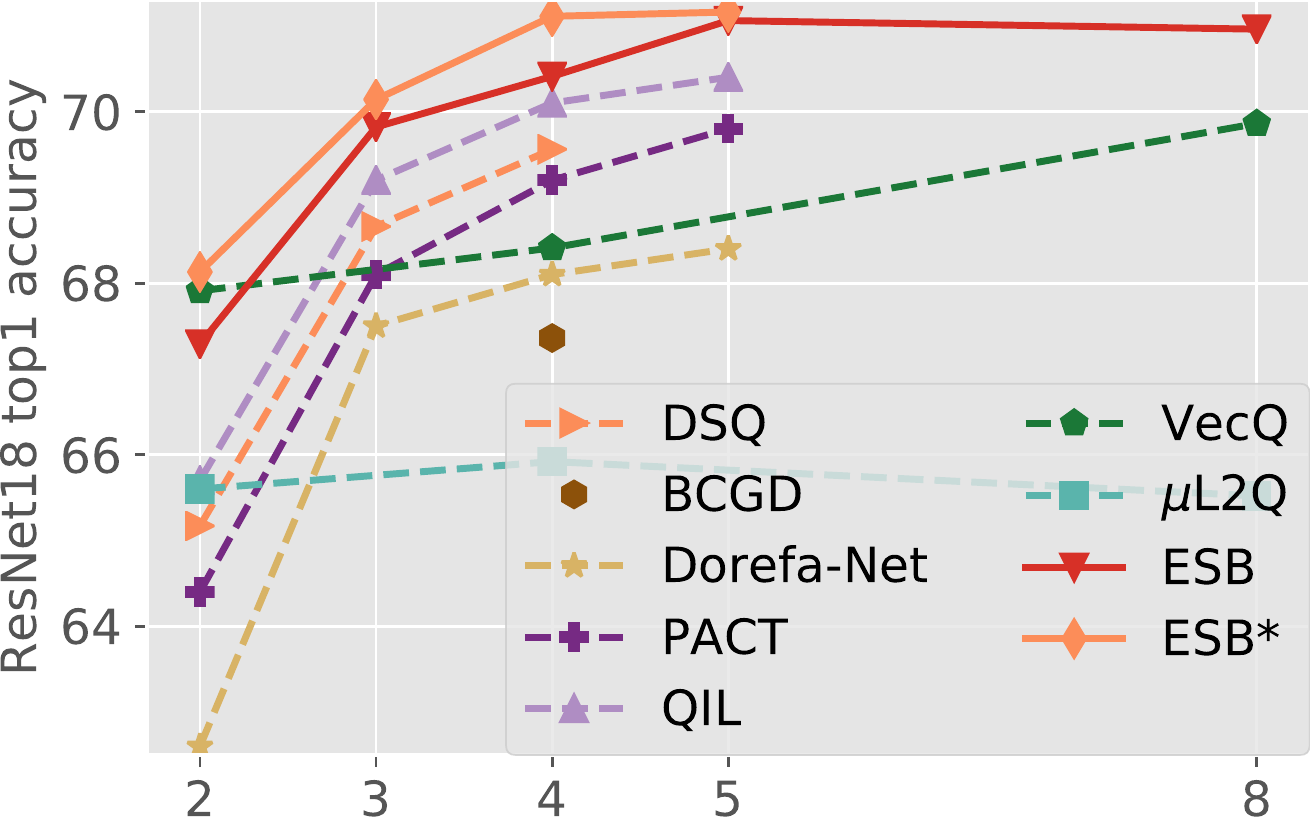}
    }\\\vspace{-8pt}
    \subfloat[\ours~vs. Nonlinear-Q on ResNet18]{
        \includegraphics[width=0.31\textwidth]{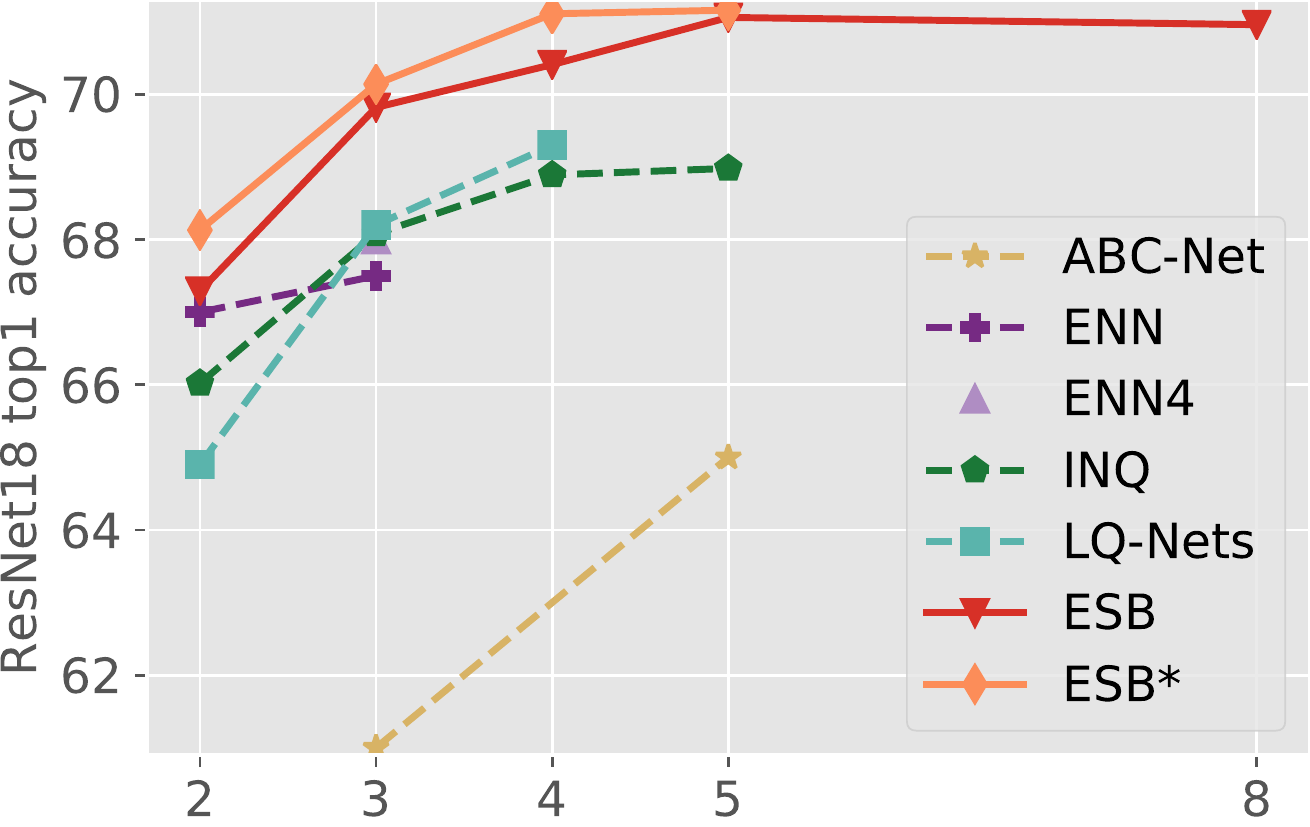}
    }
    \subfloat[\ours~vs. AutoQ and APoT on ResNet18]{
        \includegraphics[width=0.31\textwidth]{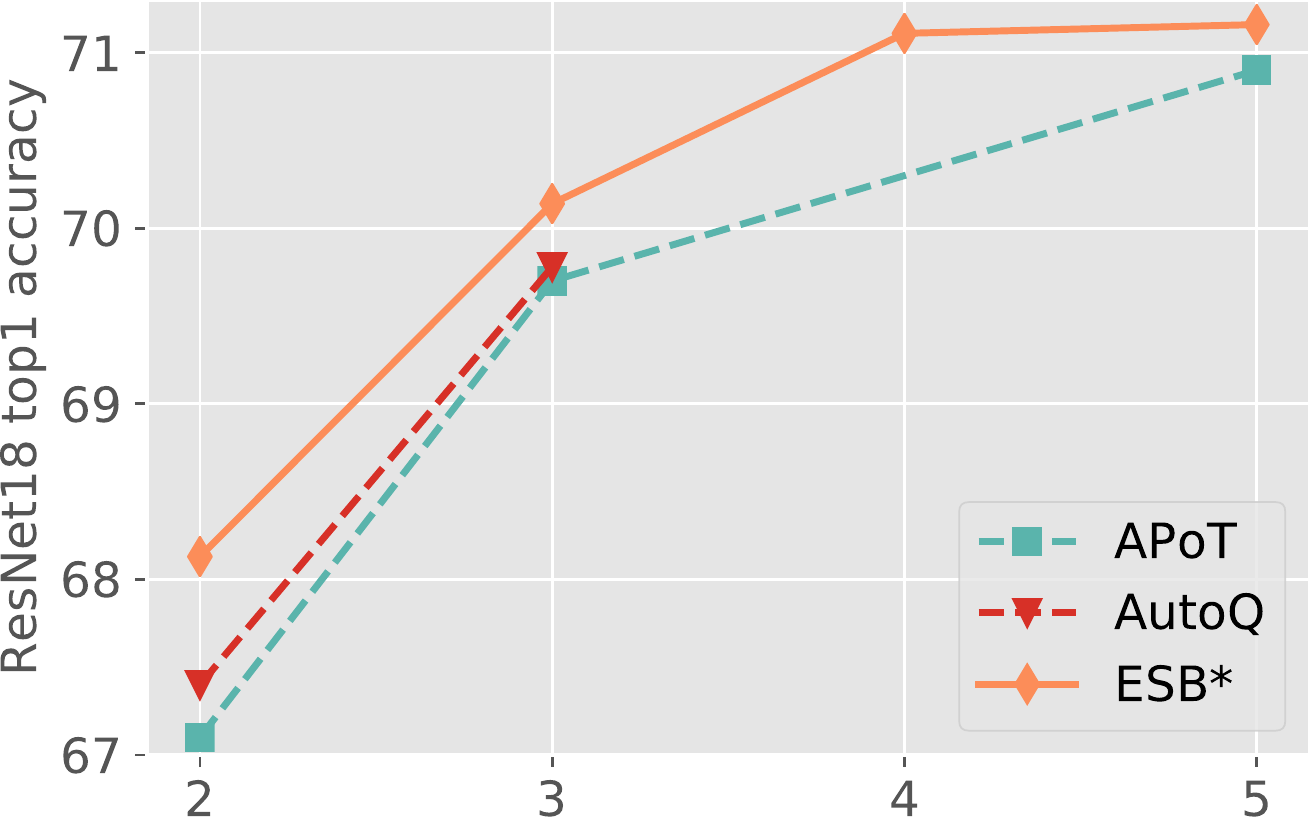}
    }
    \subfloat[MobileNetV2]{
        \includegraphics[width=0.31\textwidth]{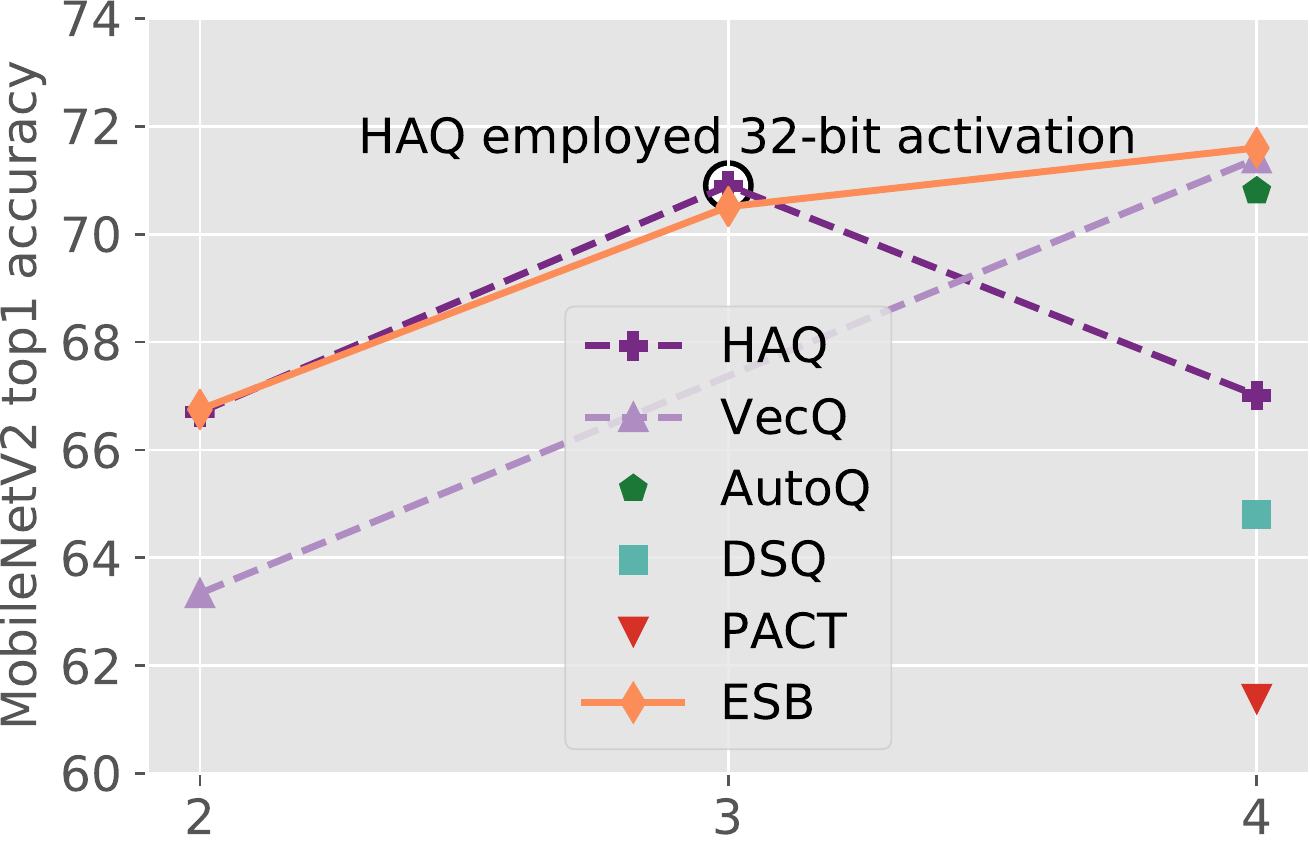}
        \label{fig:visual_accuracy_comparisons_mobilenetv2}
    }
    \vspace{-5pt}
    \caption{Visual accuracy comparisons with state-of-the-art methods on different models. Linear-Q denotes linear quantization and Nonlinear-Q refers to nonlinear quantization. The x-axis represents bitwidth.}
    \label{fig:visual_accuracy_comparisons}
    \vspace{-5pt}
\end{figure*}
\subsection{Comparison with state-of-the-art methods}
\renewcommand{\arraystretch}{1.0}
\begin{table*}[!ht]
\centering
\caption{\label{tab:comparison_works}
Comparison with the state-of-the-art approaches.}
\vspace{-7.5pt}
\setlength\tabcolsep{10.0pt} 
\begin{tabular}{l|c|c|c|c|c}
\Xhline{1.5pt} 
\multicolumn{1}{c|}{\multirow{2}{*}{{Methods}}} & \multicolumn{1}{c|}{\multirow{2}{*}{{W/A}}} & \multicolumn{2}{c|}{{AlexNet}}&
\multicolumn{2}{c}{{ResNet18}}\\\cline{3-6}
\multicolumn{1}{c|}{} & \multicolumn{1}{c|}{} &  \multicolumn{1}{c|}{{Top1}}&
\multicolumn{1}{c|}{{Top5}}&
\multicolumn{1}{c|}{{Top1}}&
\multicolumn{1}{c}{{Top5}}\\\Xhline{1.5pt}
 INQ & 2/32 & - & - & 66.02&87.13 \\
 ENN & 2/32 & 58.20&80.60 & 67.00&87.50 \\
 $\mu$L2Q & 2/32 & -&- & 65.60&86.12 \\
 VecQ & 2/8 & 58.04&80.21 &67.91&88.30 \\
 AutoQ & f2/f3 & -&- &67.40&88.18 \\
 \Xhline{0.1pt}
 TSQ & 2/2  & 58.00&80.50 & -&- \\
 DSQ & 2/2  & -&- & 65.17&- \\
 PACT & 2/2  & 55.00&77.70 & 64.40&85.60 \\
 Dorefa-Net & 2/2  & 46.40&76.80 & 62.60&84.40 \\
 LQ-Nets & 2/2  & 57.40&80.10 &64.90&85.90\\
 QIL & 2/2  & 58.10&- & 65.70&- \\
 APoT & 2/2  & -&-& 67.10&87.20\\
 \highlighted{\ours(2,0)} & 2/2 & \highlighted{58.47}&\highlighted{80.58} & \highlighted{67.29} &\highlighted{87.23} \\
 \highlighted{\ours*(2,0)} & 2/2 & \highlighted{58.73}&\highlighted{81.16} & \highlighted{68.13}& \highlighted{88.10}\\\Xhline{0.5pt}
 INQ & 5/32 & 57.39& 80.46 & 68.98&89.10 \\
 PACT & 5/5 & 55.70&77.80 & 69.80&89.30 \\
 Dorefa-Net & 5/5 & 45.10&77.90 &68.40&88.30 \\
 QIL & 5/5 & 61.90&- & 70.40&- \\
 ABC-Net & 5/5 & -&-& 65.00&85.90\\
 APoT & 5/5 & -&-& 70.90&89.70\\
\highlighted{\ours(5,2)} & 5/5 & \highlighted{62.42}&\highlighted{83.52} & \highlighted{71.06}&89.48\\
\highlighted{\ours*(5,2)} & 5/5 & \highlighted{62.89}&\highlighted{83.91} & \highlighted{71.16}&\highlighted{89.73}\\\Xhline{0.5pt}
 $\mu$L2Q & 8/32 & -&- & 65.52&86.36 \\
 VecQ & 8/8 & 61.60&83.66 & 69.86&88.90 \\
 \highlighted{\ours(8,5)} & 8/8 & \highlighted{62.94} & \highlighted{84.15} &\highlighted{70.96}&\highlighted{89.74} \\\Xhline{1.5pt}
\end{tabular}
\begin{tabular}{l|c|c|c|c|c}
\Xhline{1.5pt} \multicolumn{1}{c|}{\multirow{2}{*}{{Methods}}} & \multicolumn{1}{c|}{\multirow{2}{*}{{W/A}}} & \multicolumn{2}{c|}{{AlexNet}}&
\multicolumn{2}{c}{{ResNet18}}\\\cline{3-6}
\multicolumn{1}{c|}{} & \multicolumn{1}{c|}{} &  \multicolumn{1}{c|}{{Top1}}&
\multicolumn{1}{c|}{{Top5}}&
\multicolumn{1}{c|}{{Top1}}&
\multicolumn{1}{c}{{Top5}}\\\Xhline{1.5pt}
Referenced & 32/32 & 60.01&81.90 & 69.60&89.24 \\\Xhline{0.5pt}
 INQ & 3/32  & - &- & 68.08 & 88.36 \\
 ENN2~$\S$ & 3/32  & 59.20&81.80 & 67.50&87.90 \\
 ENN4~$\S$ & 3/32  & 60.00&82.40 & 68.00&88.30 \\
 AutoQ & f3/f4 & -&- & 69.78 & 88.38 \\
 DSQ & 3/3  & -&- & 68.66 &- \\
 PACT & 3/3  & 55.60&78.00 &68.10&88.20\\
 Dorefa-Net & 3/3  & 45.00&77.80 & 67.50&87.60 \\
 ABC-Net & 3/3  & -&- &61.00&83.20\\
 LQ-Nets & 3/3  & -&- &68.20&87.90\\
 QIL & 3/3  & 61.30&- & 69.20&- \\
 APoT & 3/3  & -&-& 69.70&88.90 \\
\highlighted{\ours(3,1)} & 3/3 & \highlighted{61.52}&\highlighted{82.96} & \highlighted{69.82}&\highlighted{89.10} \\
\highlighted{\ours*(3,1)} & 3/3 & \highlighted{61.99}&\highlighted{83.46} & \highlighted{70.14}&\highlighted{89.36}\\\Xhline{0.5pt}
 \vspace{0pt}
 INQ & 4/32& -  & - & 68.89&89.01 \\
 $\mu$L2Q & 4/32& - & - & 65.92&86.72 \\
 VecQ & 4/8 & 61.22&83.24 & 68.41&88.76 \\
 DSQ & 4/4 & -&- & 69.56&- \\
 PACT & 4/4  & 55.70&78.00 & 69.20&89.00\\  
 Dorefa-Net & 4/4  & 45.10&77.50 & 68.10&88.10\\
 LQ-Nets & 4/4  & -&- &69.30&88.80\\
 QIL & 4/4 & 62.00&- & 70.10&- \\
 BCGD & 4/4 & -&-& 67.36&87.76\\
\highlighted{\ours(4,1)} & 4/4 & \highlighted{62.19}&\highlighted{83.37} & \highlighted{70.41}&\highlighted{89.43} \\
\highlighted{\ours*(4,1)} & 4/4 & \highlighted{62.81}&\highlighted{84.02} & \highlighted{71.11}&\highlighted{89.62}\\\Xhline{1.5pt}
\end{tabular}\\\flushleft\vspace{-3pt}
    $\S$~{ENN2 indicates one-bit shift results denoted as $\{-2,+2\}$, and ENN4 represents two-bit shift denoted as $\{-4,+4\}$ in ENN \cite{ENN2017}.}
\vspace{-15pt}
\end{table*}
We then compare the \ours~with \sArt~methods under the same bit configurations.
We perform comparisons from two aspects: DNN inference efficiency and accuracy. 

For efficiency, we obtain some observations through the quantization settings, as shown in \tabref{tab:comparison_methods_config}. 
First, most of the methods, such as APoT~\cite{APoT2020}, have to reserve full precision values for the first and last layers in DNNs; otherwise, they cannot prevent decreasing accuracy. 
Methods including $\mu$L2Q~\cite{cheng2019uL2Q}, INQ~\cite{INQ2017}, and ENN~\cite{ENN2017} are forced to use 32-bit floating-point activation to maintain accuracy. 
Second, some methods~\cite{TSQ2018,ABC-Net,LQ-Nets,AutoQ} attempt to use convolutional kernel quantization to replace weight quantization, and then introduce additional full precision scaling factors. 
The number of scaling factors of these methods is related to the number of convolutional kernels, as shown in \tabref{tab:comparison_methods_config}, which can significantly increase the computing overhead.

In summary, \ours~quantizes weights and activation across all layers with elastic significant bits and only introduces one shared full precision scaling factor for all elements of weights and activation. 
In this manner, \ours~eliminates unnecessary computations and reduces the computational scale, thus improving efficiency. 
Moreover, in order to ensure a fair comparison with the methods that utilize the full precision weights/activation for the first and last layers, we perform additional experiments without quantizing the first and last layers. 
The results are denoted as \highlighted{\ours*}.

\renewcommand{\arraystretch}{1.0}
\begin{table*}[!t]
\centering
\caption{\label{tab:results_mobilenetv2}
Results on light-weight MobileNetV2~\cite{sandler2018mobilenetv2}.}
\vspace{-7.5pt}
\setlength\tabcolsep{20.5pt} 
\begin{tabular}{l|c|c|c}
\Xhline{1.5pt} 
\multicolumn{1}{c|}{\multirow{2}{*}{{Methods}}} & \multicolumn{1}{c|}{\multirow{2}{*}{{W/A}}} & \multicolumn{2}{c}{{MobileNetV2}}\\\cline{3-4}
\multicolumn{1}{c|}{} & \multicolumn{1}{c|}{} &  \multicolumn{1}{c|}{{Top1}}&
\multicolumn{1}{c}{{Top5}}\\\Xhline{1.5pt}
 Referenced & 32/32 &  71.30&90.10  \\\Xhline{0.5pt}
HAQ & f2/32& 66.75&87.32 \\
VecQ & 2/8 & 63.34 &84.42\\
{\ours}(2,0) & {2/2} & \highlighted{66.75}&87.14 \\\Xhline{0.5pt}
HAQ & f3/32& \highlighted{70.90}&89.76 \\
{\ours}(3,1) & {3/3} & 70.51&89.46 \\\Xhline{1.5pt}
\end{tabular}
\begin{tabular}{l|c|c|c}
\Xhline{1.5pt} 
\multicolumn{1}{c|}{\multirow{2}{*}{{Methods}}} & \multicolumn{1}{c|}{\multirow{2}{*}{{W/A}}} & \multicolumn{2}{c}{{MobileNetV2}}\\\cline{3-4}
\multicolumn{1}{c|}{} & \multicolumn{1}{c|}{} &  \multicolumn{1}{c|}{{Top1}}&
\multicolumn{1}{c}{{Top5}}\\\Xhline{1.5pt}
 VecQ & 4/8 & 71.40&90.41 \\
 HAQ & f4/f4  & 67.01&87.46 \\
 AutoQ & f4/f4  & 70.80&90.33 \\
 DSQ & 4/4  & 64.80&-\\
PACT & 4/4  & 61.39&83.72\vspace{1pt}\\
 {\ours}(4,1) & {4/4} & \highlighted{71.60}&90.23  \\\Xhline{1.5pt}
\end{tabular}\flushleft\vspace{-5pt}
\footnotemark[1]{The results for Referenced refer to~\cite{VecQ}, and the results of PACT are cited from~\cite{wang2019haq}.}
\vspace{-5pt}
\end{table*}
\begin{table*}[!ht]
    \centering
    \caption{\ours~accelerator resource utilization and performance based on AlexNet.}
    \label{tab:hardware_implementation}
    \vspace{-7.5pt}
    \setlength\tabcolsep{7.2pt} 
    \begin{tabular}{c|c|ccc|cccc|cc|c|c|c}\Xhline{1.5pt}
\multirow{2}{*}{\ours($b,k$)} &\multirow{2}{*}{Alias} &
  \multicolumn{3}{c|}{MAC (\#LUTs)} &
  \multicolumn{4}{c|}{Resource Utilization (\%)} &
  \multicolumn{2}{c|}{Concurrency} &
  Clock &
  Power &
  \multirow{2}{*}{GOPS} \\\cline{3-11}
& & Mul.$\dagger$ & Acc.$\dagger$ & Mul.+Acc. & LUTs  & FF    & BRAM  & DSP  & Tn & Tm  & (MHz) & (W)  &         \\\Xhline{1.5pt}
\ours(2,0) & Ternary & \highlighted{2}  & \highlighted{12} & \highlighted{14} & \highlighted{63.94} & 16.27 & 47.64 & 4.25 & 32 & 200 & 145 & 3.57 & 1856.00 \\\Xhline{0.5pt}
\ours(3,0) & PoT & 6  & 16 & 22  & 69.81 & 18.21 & 41.06 & 4.48 & 32 & 160 & 145 & 4.30 & 1484.80 \\
\ours(3,1) & Fixed$\dagger$& \highlighted{5}  & \highlighted{15} & \highlighted{20} & \highlighted{62.78} & 17.89 & 41.06 & 4.48 & 32 & 160 & 145 & 4.67 & 1484.80 \\\Xhline{0.5pt}
\ours(4,0) & PoT&\highlighted{11} & 24 &35 & 73.24 & 18.04 & 46.33 & 4.25 & 32 & 96  & 145 & 3.67 & 890.88  \\
\ours(4,1) & \ours &\highlighted{11} & 19 &\highlighted{30}  & \highlighted{61.71} & 16.33 & 30.54 & 4.25 & 32 & 96  & 145 & 4.08 & 890.88  \\
\ours(4,2) &Fixed & 19 & \highlighted{17} & 36  & 70.86 & 15.21 & 30.54 & 4.25 & 32 & 96  & 145 & 4.02 & 890.88  \\\Xhline{0.5pt}
\ours(5,1) & \ours& \highlighted{20} & 27 & \highlighted{47} & \highlighted{60.37} & 15.99 & 49.84 & 4.33 & 32 & 64  & 145 & 3.93 & 593.92  \\
\ours(5,2) & \ours& 26 & 21 & 47 & 60.95 & 14.44 & 49.84 & 4.33 & 32 & 64  & 145 & 3.89 & 593.92  \\
\ours(5,3) & Fixed& 41 & 19&60 & 66.87 & 14.17 & 39.31 & 4.33 & 32 & 64  & 145 & 3.79 & 593.92  \\\Xhline{0.5pt}
\ours(6,2)& \ours & \highlighted{36} &29 &\highlighted{65} & 61.07 & 14.68 & 45.01 & 4.68 & 16 & 92  & 145 & 3.70 & 426.88  \\
\ours(6,3)& \ours & 45 & 23 &68 & \highlighted{57.64} & 13.89 & 45.01 & 4.68 & 16 & 92  & 145 & 4.78 & 426.88  \\
\ours(6,4)& Fixed & 55 &\highlighted{21} &76 & 65.11 & 13.67 & 45.01 & 4.68 & 16 & 92  & 145 & 4.65 & 426.88  \\\Xhline{0.5pt}
\ours(7,3)& \ours & 59 & 31 & 90 & 85.24 & 14.99 & 41.06 & 4.44 & 16 & 80  & 145 & 4.02 & 371.20  \\
\ours(7,4)& \ours & \highlighted{57} & 25 &\highlighted{82} & 60.88 & 13.86 & 41.06 & 4.44 & 16 & 80  & 145 & 3.73 & 371.20  \\
\ours(7,5)& Fixed & 66 & \highlighted{23} & 89 & \highlighted{59.78} & 14.19 & 41.06 & 4.44 & 16 & 80  & 145 & 4.12 & 371.20  \\\Xhline{0.5pt}
\ours(8,4)& \ours & 71 & 33& 104 & 74.90 & 15.49 & 37.77 & 4.44 & 16 & 70  & 145 & 4.89 & 324.80  \\
\ours(8,5)& \ours & \highlighted{69} & 27& \highlighted{96} & \highlighted{59.57} & 14.55 & 37.77 & 4.44 & 16 & 70  & 145 & 3.99 & 324.80  \\
\ours(8,6)& Fixed & 86 & \highlighted{25} & 111 & 63.17 & 14.35 & 37.77 & 4.44 & 16 & 70  & 145 & 3.61 & 324.80  \\\Xhline{1.5pt}
\end{tabular}\\\flushleft
\vspace{-5pt}
$\dagger$ Mul. denotes the multiplier; Acc. represents the accumulator; Fixed refers to the fixed-point format of linear quantization.
\vspace{-15pt}
\end{table*}
For accuracy comparison, we survey and replicate abundant existing methods as our comparison targets and list all the quantitative results in \tabref{tab:comparison_works}.
Besides, to highlight the accuracy improvements of ESB, we provide the accuracy comparison by different quantization categories (Linear-Q and Nonlinear-Q) in \figref{fig:visual_accuracy_comparisons}.
\ours~consistently outperforms the \sArt~approaches across all given bits and improves the top 1 results of AlexNet and ResNet18 by \alexnetOimp~and \resnetOimp, respectively,~on average. 
Even by quantizing the weights and activation for the first and last layers, \ours~still outperforms APoT by $\highlighted{0.16\%}$ for accuracy on average. 
APoT performs a simple sum of several PoTs, which leads to the deviation of the target bell-shaped distribution far away. 
\ours~also has higher accuracy compared with the approaches using full precision activation such as INQ and ENN, with improvements of $\highlighted{1.65\%}$ and $\highlighted{1.48\%}$ over ResNet18, respectively. 
In addition, \ours~is suitable for 2-bit quantization and is better than 2-bit schemes such as TSQ, DSQ, and APoT in terms of accuracy.
Furthermore, \ours* consistently achieves the best accuracy and outperforms all \sArt~works over AlexNet and ResNet18, respectively. 
Even under two given bits, compared with APoT, \ours*~can improve accuracy by up to $\highlighted{1.03\%}$. 
On average, \ours* outperforms APoT by $\highlighted{0.58\%}$ in terms of accuracy.

The reason \ours~can achieve high accuracy is that \ours~fits the distribution well with the \ours~and DDA.
Significant bits provide a fine-grain control of the density of quantized values for various distributions of DNN weights/activation. Furthermore, DDA can minimize the difference between the \ours~quantized value distribution and the original distribution to facilitate \ours~to seek the best distribution.
Consequently, fine control granularity inspires \ours~to achieve accuracy gains. 

\vspace{-4pt}
\subsection{\ours~on light-weight model}
We next evaluate the \ours~on the popular light-weight model,
MobileNetV2~\cite{sandler2018mobilenetv2}, to verify the superiority of \ours. The important results of the \ours~are presented in \tabref{tab:results_mobilenetv2} and \figref{fig:visual_accuracy_comparisons_mobilenetv2}. 
We would like to emphasize that the \ours~quantizes the weights and activation for all layers of MobileNetV2. 
\ours~outperforms DSQ~\cite{DSQ} and VecQ~\cite{VecQ}, and achieves an average accuracy improvement of~\mobileOimp~compared with \sArt~methods. 
Lowering to 2 bits for quantization, \ours~can maintain an accuracy of $\highlighted{66.75\%}$, which is similar to the accuracy of HAQ with mixed-precision weights and full precision activation.
Experiments on MobileNetV2 demonstrate that \ours~quantization is an effective strategy to fit the distributions of weights or activation by controlling the number of significant bits of quantized values, so as to mitigate the accuracy degradation of low-bitwidth quantization.

\subsection{Hardware evaluation}
In this subsection, we implement \ours~accelerator on the FPGA platform to verify the hardware benefits of the \ours. 
The objectives of the hardware evaluation are twofold: 
1) exploring the hardware resource utilization of \ours\ with different configurations;
2) providing insight into energy efficiency and presenting comparisons among \ours~accelerator, \sArt~designs, and CPU/GPU platforms.

\cusparagraph{Resource utilization.}
In general, there are limited computing resources such as LUTs, FFs, BRAMs, and DSPs in FPGAs. 
Hardware implementations usually need to fully utilize the resources to achieve a high concurrency and clock frequency.
In our experiments, we freeze the clock frequency to 145 MHz.
For a fair comparison, we employ the LUT instead of the DSP to implement fraction multiplication.
The resource utilization and concurrency under various \ours\ implementations are listed in \tabref{tab:hardware_implementation}.

First, we show the effect of how significant bits affect the MAC, including the multiplier and accumulator, at the bit-wise level.
We observe that fewer significant bits of fraction, that is, a smaller value of $k$, can reduce the resource consumption of the multiplier.
With the same number of given bits, the consumption of LUTs for one multiplier can decrease by 6.3 on an average if $k$ is decreased by 1.
This is because fewer significant bits can reduce the consumption for fraction-multiplication, that is, $(\zeta_i+0.f_i)*(\zeta_j+0.f_j)$ in \eqref{eq:esb_multiplication}, thus consuming fewer computing resources.
Nevertheless, the results also show that decreasing $k$ can increase the resource consumption of the accumulator.
Using the same number of given bits, the consumption of LUTs of the accumulator increases by 3.6 on average, if $k$ is decreased by 1.
This is because the relatively large exponents, that is, $e=e_i\zeta_i+e_j\zeta_j$ in \eqref{eq:esb_multiplication}, require a high output bitwidth and thus increase the resource budgets of the accumulator. 
In terms of MAC consumption, \ours~can strike a better balance between the multiplier and accumulator, utilizing fewer resources than fixed-point (Fixed) and PoT.

\begin{figure}[!t]
    \centering
    \includegraphics[width=1\columnwidth]{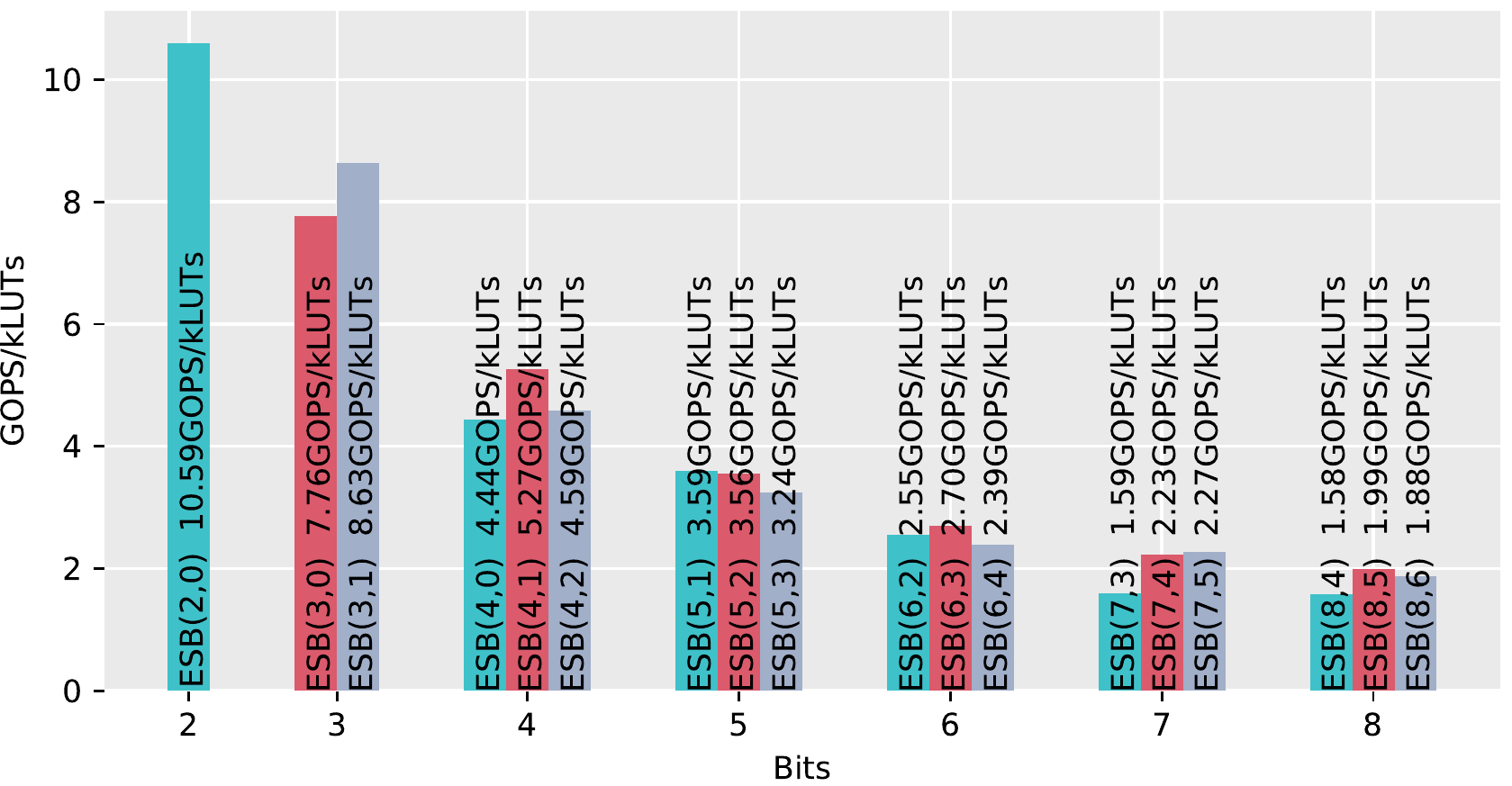}
    \vspace{-20pt}
    \caption{Peak performance (GOPS) per 1k LUTs.}
    \label{fig:utilization-performance}
    \vspace{-20pt}
\end{figure}
Second, we verify that the elastic significant bits can improve the performance of the accelerator design. We take the achieved performance under per 1k LUTs for a quantitative evaluation. 
As shown in \figref{fig:utilization-performance}, the accelerator of \ours(2,0) exhibits the best performance with 10.59 GOPS under 1k LUTs, because it costs only 14 LUTs for MAC.
For the other cases, the best results are 8.63, 5.27, 3.59, 2.70, 2.27, and 1.99 GOPS/kLUTs for configurations (3,1), (4,1), (5,1), (6,3), (7,5), and (8-5) across various quantizing bits, and the improvements are up to \highlighted{11.19\%}, \highlighted{18.68\%}, \highlighted{10.77\%}, \highlighted{12.96\%}, \highlighted{42.59\%}, and \highlighted{25.73\%} compared with the results of (3,0), (4,0), (5,3), (6,4), (7,3), and (8,4), respectively. By employing desirable number of significant bits in accelerator design, the performance within per 1k LUTs can improve by \highlighted{20.32\%} on average without requiring more given bits. 

\begin{figure}[!t]
    \centering
    \includegraphics[width=1\columnwidth]{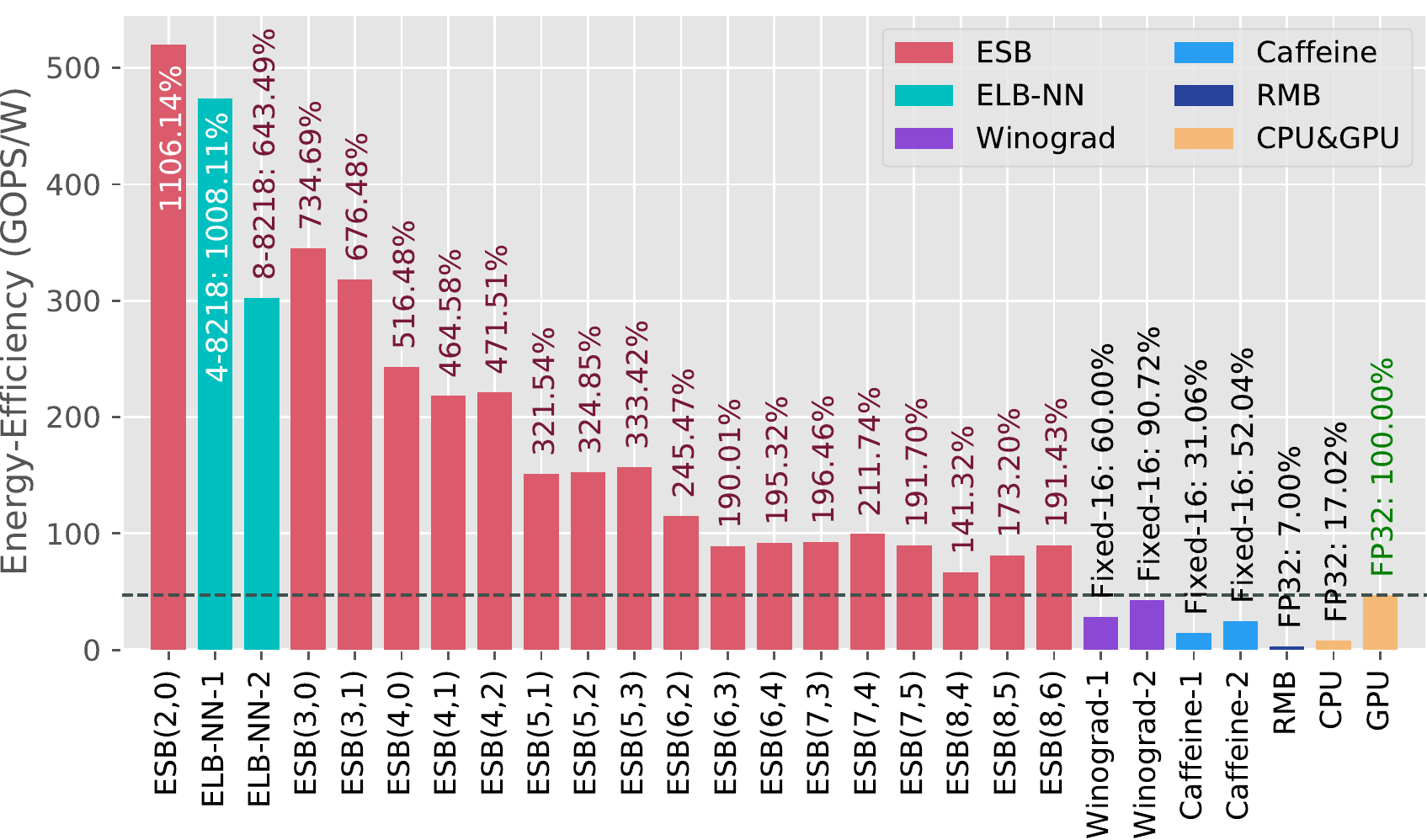}
    \vspace{-20pt}
    \caption{Energy efficiency comparison among \sArt~designs and CPU/GPU.} 
    \label{fig:energy-efficiency-cpu-gpu}
    \vspace{-17pt}
\end{figure}

\cusparagraph{Energy efficiency.}
Energy efficiency is a critical criterion for evaluating a computing system, which is defined as the number of operations that can be completed within a unit of energy (J)~\cite{FPGA_survey_Guo2019}. Here, we define the energy efficiency as the ratio of peak performance (GOPS) to power (W). 
\figref{fig:energy-efficiency-cpu-gpu} presents
the energy efficiency comparisons of our accelerator with the recent FPGA designs, including ELB-NN\cite{Design-flow-of-accelerating}, Winograd~\cite{winograd}, Caffeine~\cite{caffeine}, and Roofline-model-based method (RMB)~\cite{Optimizing-FPGA-based-accelerator}, and CPU (Xeon E5-2678 v3)/GPU (RTX 2080 Ti) platforms.

Compared with the ELB-NN~\cite{Design-flow-of-accelerating} for the formats of 8-8218 and 4-8218 (extremely low bitwidth quantization and the least bit width is 1), \ours(2,0) achieves higher energy efficiency, with improvements of \highlighted{71.90\%} and \highlighted{9.73\%}, respectively. 
Compared with the accelerators using high-bitwidth formats, such as Fixed-8~\cite{Design-flow-of-accelerating,lite-cnn} and Fixed-16~\cite{winograd,caffeine}, or even float-point data type (FP32)~\cite{Optimizing-FPGA-based-accelerator}, \ours(8,6) can achieve a higher energy efficiency, and the improvement is by \highlighted{8$\times$} on average, and up to \highlighted{26$\times$} (\ours(8,6) vs. \cite{Optimizing-FPGA-based-accelerator}).
\ours~accelerator also achieves a higher energy efficiency than CPU/GPU platforms across all bitwidths and configurations.
Their energy efficiency increased by \highlighted{22$\times$} and \highlighted{4$\times$} on average compared with CPU and GPU, and up to \highlighted{65$\times$} and \highlighted{11$\times$} improvement at 2-bit, respectively.
In addition, \ours~accelerator can be implemented with various bitwidths and configurations, which can help us achieve a fine-grained trade-off between inference accuracy and computing efficiency. For example, \ours(3,0) is a reasonable choice under a tolerated accuracy of 60\% (AlexNet Top1), which can achieve an energy efficiency of 345.30GOPS/W with an accuracy of 60.73\%. \ours(3,1) with 317.94GOPS/W is a reasonable choice under a tolerated accuracy of 61\%, while \ours(4,1) with 218.35GOPS/W is the desirable choice under a tolerated accuracy of 62\%.
\section{Conclusion}\label{sec:conclusion}
In this paper, we propose an \ours\ for DNN quantization to strike a good balance between accuracy and efficiency. 
We explore a new perspective to improve quantization accuracy and concurrently decrease computing resources by controlling significant bits of quantized values. 
We propose the mathematical expression of \ours~to unify various distributions of quantized values. The unified expression outlines the linear quantization and PoT quantization, and guides DNN compression from both accuracy and efficiency perspectives by fitting distribution using \ours.
Moreover, we design an \ours~projection operator involving shifting and rounding operations, which are cheap bit operations at the hardware level without complex conversions. 
Fewer significant bits representing weight/activation and cheap projection of quantization are what hardware desires because they meet hardware computing natures.  
In addition, the proposed DDA always finds a desirable distribution for \ours~quantization to reduce accuracy degradation. 

Experimental results show that \ours~quantization outperforms \sArt~methods and achieves higher accuracy across all quantizing bitwidths.
The hardware implementation also demonstrates that quantizing DNNs with elastic significant bits can achieve higher energy efficiency without increasing the quantizing bitwidth.
In the future, we will further study the real performance benefits of our \ours~on edge devices.
\section{Acknowledgments}
This work is partially supported by the National Key Research and Development Program of China (2018YFB2100300), 
the National Natural Science Foundation (62002175), 
the Natural Science Foundation of Tianjin (19JCQNJC00600),
the State Key Laboratory of Computer Architecture (ICT,CAS) under Grant No. CARCHB202016, CARCH201905, 
and the Innovation Fund of Chinese Universities Industry-University-Research (2020HYA01003).

\bibliographystyle{IEEEtran}
\vspace{-5pt}
\bibliography{main}

\begin{thebibliography}{10}
\providecommand{\url}[1]{#1}
\csname url@samestyle\endcsname
\providecommand{\newblock}{\relax}
\providecommand{\bibinfo}[2]{#2}
\providecommand{\BIBentrySTDinterwordspacing}{\spaceskip=0pt\relax}
\providecommand{\BIBentryALTinterwordstretchfactor}{4}
\providecommand{\BIBentryALTinterwordspacing}{\spaceskip=\fontdimen2\font plus
\BIBentryALTinterwordstretchfactor\fontdimen3\font minus
  \fontdimen4\font\relax}
\providecommand{\BIBforeignlanguage}[2]{{%
\expandafter\ifx\csname l@#1\endcsname\relax
\typeout{** WARNING: IEEEtran.bst: No hyphenation pattern has been}%
\typeout{** loaded for the language `#1'. Using the pattern for}%
\typeout{** the default language instead.}%
\else
\language=\csname l@#1\endcsname
\fi
#2}}
\providecommand{\BIBdecl}{\relax}
\BIBdecl

\bibitem{he2016resnet}
K.~He, X.~Zhang, S.~Ren, and J.~Sun, ``Deep residual learning for image
  recognition,'' in \emph{{IEEE CVPR}}.\hskip 1em plus 0.5em minus 0.4em\relax
  {IEEE} Computer Society, 2016, pp. 770--778.

\bibitem{alexnet}
A.~Krizhevsky, I.~Sutskever, and G.~E. Hinton, ``Imagenet classification with
  deep convolutional neural networks,'' in \emph{{NIPS}}, 2012, pp. 1106--1114.

\bibitem{PatDNN}
W.~Niu, X.~Ma, S.~Lin, S.~Wang, X.~Qian, X.~Lin, Y.~Wang, and B.~Ren, ``Patdnn:
  Achieving real-time dnn execution on mobile devices with pattern-based weight
  pruning,'' in \emph{ASPLOS}.\hskip 1em plus 0.5em minus 0.4em\relax
  Association for Computing Machinery, 2020, p. 907–922.

\bibitem{rastegari2016xnor}
M.~Rastegari, V.~Ordonez, J.~Redmon, and A.~Farhadi, ``Xnor-net: Imagenet
  classification using binary convolutional neural networks,'' in
  \emph{{ECCV}}, B.~Leibe, J.~Matas, N.~Sebe, and M.~Welling, Eds., vol.
  9908.\hskip 1em plus 0.5em minus 0.4em\relax Springer, 2016, pp. 525--542.

\bibitem{chen2019tdla}
Y.~{Chen}, K.~{Zhang}, C.~{Gong}, C.~{Hao}, X.~{Zhang}, T.~{Li}, and D.~{Chen},
  ``{T-DLA}: An open-source deep learning accelerator for ternarized dnn models
  on embedded fpga,'' in \emph{ISVLSI}, 2019, pp. 13--18.

\bibitem{zhou2016dorefa}
S.~Zhou, Z.~Ni, X.~Zhou, H.~Wen, Y.~Wu, and Y.~Zou, ``Dorefa-net: Training low
  bitwidth convolutional neural networks with low bitwidth gradients,''
  \emph{CoRR}, vol. abs/1606.06160, 2016.

\bibitem{DSQ}
R.~Gong, X.~Liu, S.~Jiang, T.~Li, P.~Hu, J.~Lin, F.~Yu, and J.~Yan,
  ``Differentiable soft quantization: Bridging full-precision and low-bit
  neural networks,'' in \emph{{IEEE ICCV}}.\hskip 1em plus 0.5em minus
  0.4em\relax {IEEE}, 2019, pp. 4851--4860.

\bibitem{cheng2019uL2Q}
C.~Gong, T.~Li, Y.~Lu, C.~Hao, X.~Zhang, D.~Chen, and Y.~Chen,
  ``{\(\mathrm{\mu}\)}l2q: An ultra-low loss quantization method for {DNN}
  compression,'' in \emph{IJCNN}.\hskip 1em plus 0.5em minus 0.4em\relax
  {IEEE}, 2019, pp. 1--8.

\bibitem{VecQ}
C.~Gong, Y.~Chen, Y.~Lu, T.~Li, C.~Hao, and D.~Chen, ``Vecq: Minimal loss dnn
  model compression with vectorized weight quantization,'' \emph{IEEE
  Transactions on Computers}, vol.~70, no.~5, pp. 696--710, 2021.

\bibitem{PostQ}
R.~Banner, Y.~Nahshan, and D.~Soudry, ``Post training 4-bit quantization of
  convolutional networks for rapid-deployment,'' in \emph{{NIPS}},
  vol.~32.\hskip 1em plus 0.5em minus 0.4em\relax Curran Associates, Inc.,
  2019.

\bibitem{SYQ}
J.~Faraone, N.~J. Fraser, M.~Blott, and P.~H.~W. Leong, ``{SYQ:} learning
  symmetric quantization for efficient deep neural networks,'' in \emph{{IEEE
  CVPR}}.\hskip 1em plus 0.5em minus 0.4em\relax Computer Vision Foundation /
  {IEEE} Computer Society, 2018, pp. 4300--4309.

\bibitem{QAT}
B.~Jacob, S.~Kligys, B.~Chen, M.~Zhu, M.~Tang, A.~Howard, H.~Adam, and
  D.~Kalenichenko, ``Quantization and training of neural networks for efficient
  integer-arithmetic-only inference,'' in \emph{{IEEE CVPR}}, 2018, pp.
  2704--2713.

\bibitem{LogQ}
D.~Miyashita, E.~H. Lee, and B.~Murmann, ``Convolutional neural networks using
  logarithmic data representation,'' \emph{CoRR}, vol. abs/1603.01025, 2016.

\bibitem{AddNet}
J.~Faraone, M.~Kumm, M.~Hardieck, P.~Zipf, X.~Liu, D.~Boland, and P.~H.~W.
  Leong, ``Addnet: Deep neural networks using fpga-optimized multipliers,''
  \emph{IEEE T VLSI SYST}, vol.~28, no.~1, pp. 115--128, 2020.

\bibitem{APoT2020}
Y.~Li, X.~Dong, and W.~Wang, ``Additive powers-of-two quantization: An
  efficient non-uniform discretization for neural networks,'' in \emph{ICLR
  2020}, 2020.

\bibitem{han2015deep}
S.~Han, H.~Mao, and W.~J. Dally, ``Deep compression: Compressing deep neural
  network with pruning, trained quantization and huffman coding,'' in
  \emph{ICLR}, Y.~Bengio and Y.~LeCun, Eds., 2016.

\bibitem{INQ2017}
A.~Zhou, A.~Yao, Y.~Guo, L.~Xu, and Y.~Chen, ``Incremental network
  quantization: Towards lossless cnns with low-precision weights,''
  \emph{CoRR}, vol. abs/1702.03044, 2017.

\bibitem{ENN2017}
C.~Leng, Z.~Dou, H.~Li, S.~Zhu, and R.~Jin, ``Extremely low bit neural network:
  Squeeze the last bit out with {ADMM},'' in \emph{{AAAI}}, S.~A. McIlraith and
  K.~Q. Weinberger, Eds.\hskip 1em plus 0.5em minus 0.4em\relax {AAAI} Press,
  2018, pp. 3466--3473.

\bibitem{QIL}
S.~Jung, C.~Son, S.~Lee, J.~Son, J.~Han, Y.~Kwak, S.~J. Hwang, and C.~Choi,
  ``Learning to quantize deep networks by optimizing quantization intervals
  with task loss,'' in \emph{{IEEE CVPR}}.\hskip 1em plus 0.5em minus
  0.4em\relax Computer Vision Foundation / {IEEE}, 2019, pp. 4350--4359.

\bibitem{BCGD}
P.~Yin, S.~Zhang, J.~Lyu, S.~J. Osher, Y.~Qi, and J.~Xin, ``Blended coarse
  gradient descent for full quantization of deep neural networks,''
  \emph{CoRR}, vol. abs/1808.05240, 2018.

\bibitem{ABC-Net}
X.~Lin, C.~Zhao, and W.~Pan, ``Towards accurate binary convolutional neural
  network,'' in \emph{{NIPS}}, I.~Guyon, U.~von Luxburg, S.~Bengio, H.~M.
  Wallach, R.~Fergus, S.~V.~N. Vishwanathan, and R.~Garnett, Eds., 2017, pp.
  345--353.

\bibitem{CNN-for-skin-lesions}
X.~Fan, M.~Dai, C.~Liu, F.~Wu, X.~Yan, Y.~Feng, Y.~Feng, and B.~Su, ``Effect of
  image noise on the classification of skin lesions using deep convolutional
  neural networks,'' \emph{Tsinghua Science and Technology}, vol.~25, no.~3,
  pp. 425--434, 2020.

\bibitem{DL-based-polar-emotion-classification}
Q.~Cao, W.~Zhang, and Y.~Zhu, ``Deep learning-based classification of the polar
  emotions of "moe"-style cartoon pictures,'' \emph{Tsinghua Science and
  Technology}, vol.~26, no.~3, pp. 275--286, 2021.

\bibitem{DL-bigdatama}
W.~Zhong, N.~Yu, and C.~Ai, ``Applying big data based deep learning system to
  intrusion detection,'' \emph{Big Data Min. Anal.}, vol.~3, no.~3, pp.
  181--195, 2020.

\bibitem{DRL-for-MRN}
Z.~Kai and Z.~Tao, ``Deep reinforcement learning based mobile robot navigation:
  A review,'' \emph{Tsinghua Science and Technology}, vol.~26, no.~5, pp.
  674--691, 2021.

\bibitem{TSQ2018}
P.~Wang, Q.~Hu, Y.~Zhang, C.~Zhang, Y.~Liu, and J.~Cheng, ``Two-step
  quantization for low-bit neural networks,'' in \emph{{IEEE CVPR}}, 2018, pp.
  4376--4384.

\bibitem{PACT}
J.~Choi, Z.~Wang, S.~Venkataramani, P.~I. Chuang, V.~Srinivasan, and
  K.~Gopalakrishnan, ``{PACT:} parameterized clipping activation for quantized
  neural networks,'' \emph{CoRR}, vol. abs/1805.06085, 2018.

\bibitem{wang2019haq}
K.~Wang, Z.~Liu, Y.~Lin, J.~Lin, and S.~Han, ``Haq: Hardware-aware automated
  quantization with mixed precision,'' in \emph{{IEEE CVPR}}, 2019, pp.
  8612--8620.

\bibitem{ghasemzadeh2018rebnet}
M.~Ghasemzadeh, M.~Samragh, and F.~Koushanfar, ``Rebnet: Residual binarized
  neural network,'' in \emph{26th {IEEE} Annual International Symposium on
  Field-Programmable Custom Computing Machines, {FCCM} 2018, Boulder, CO, USA,
  April 29 - May 1, 2018}.\hskip 1em plus 0.5em minus 0.4em\relax {IEEE}
  Computer Society, 2018, pp. 57--64.

\bibitem{LQ-Nets}
D.~Zhang, J.~Yang, D.~Ye, and G.~Hua, ``Lq-nets: Learned quantization for
  highly accurate and compact deep neural networks,'' in \emph{{ECCV}}, ser.
  Lecture Notes in Computer Science, V.~Ferrari, M.~Hebert, C.~Sminchisescu,
  and Y.~Weiss, Eds., vol. 11212.\hskip 1em plus 0.5em minus 0.4em\relax
  Springer, 2018, pp. 373--390.

\bibitem{AutoQ}
Q.~Lou, F.~Guo, M.~Kim, L.~Liu, and L.~Jiang., ``Autoq: Automated kernel-wise
  neural network quantization,'' in \emph{ICLR}, 2020.

\bibitem{BFP}
M.~Drumond, T.~Lin, M.~Jaggi, and B.~Falsafi, ``End-to-end dnn training with
  block floating point arithmetic,'' \emph{CoRR}, vol. abs/1804.01526, 2018.

\bibitem{Flexpoint}
U.~K{\"o}ster, T.~Webb, X.~Wang, M.~Nassar, A.~K. Bansal, W.~Constable,
  O.~Elibol, S.~Gray, S.~Hall, L.~Hornof \emph{et~al.}, ``Flexpoint: An
  adaptive numerical format for efficient training of deep neural networks,''
  in \emph{{NIPS}}, 2017, pp. 1742--1752.

\bibitem{PositArithmetic}
J.~Gustafson and I.~Yonemoto, ``Beating floating point at its own game: Posit
  arithmetic,'' \emph{Supercomputing Frontiers and Innovations}, vol.~4, no.~2,
  2017.

\bibitem{DeepPositron}
Z.~{Carmichael}, H.~F. {Langroudi}, C.~{Khazanov}, J.~{Lillie}, J.~L.
  {Gustafson}, and D.~{Kudithipudi}, ``Deep positron: A deep neural network
  using the posit number system,'' in \emph{DATE}, 2019, pp. 1421--1426.

\bibitem{AdaptivFloat}
T.~Tambe, E.~Yang, Z.~Wan, Y.~Deng, V.~J. Reddi, A.~M. Rush, D.~Brooks, and
  G.~Wei, ``Adaptivfloat: {A} floating-point based data type for resilient deep
  learning inference,'' \emph{CoRR}, vol. abs/1909.13271, 2019.

\bibitem{hashnet}
W.~Chen, J.~Wilson, S.~Tyree, K.~Weinberger, and Y.~Chen, ``Compressing neural
  networks with the hashing trick,'' in \emph{ICML}.\hskip 1em plus 0.5em minus
  0.4em\relax PMLR, 2015, pp. 2285--2294.

\bibitem{ioffe2015batch}
S.~Ioffe and C.~Szegedy, ``Batch normalization: Accelerating deep network
  training by reducing internal covariate shift,'' in \emph{{ICML}}, F.~R. Bach
  and D.~M. Blei, Eds., vol.~37.\hskip 1em plus 0.5em minus 0.4em\relax
  JMLR.org, 2015, pp. 448--456.

\bibitem{lee2015deeply}
C.~Lee, S.~Xie, P.~W. Gallagher, Z.~Zhang, and Z.~Tu, ``Deeply-supervised
  nets,'' in \emph{AISTATS}, ser. {JMLR} Workshop and Conference Proceedings,
  vol.~38.\hskip 1em plus 0.5em minus 0.4em\relax JMLR.org, 2015.

\bibitem{sandler2018mobilenetv2}
M.~Sandler, A.~G. Howard, M.~Zhu, A.~Zhmoginov, and L.~Chen, ``Mobilenetv2:
  Inverted residuals and linear bottlenecks,'' in \emph{{IEEE CVPR}}.\hskip 1em
  plus 0.5em minus 0.4em\relax {IEEE} Computer Society, 2018, pp. 4510--4520.

\bibitem{deng2009imagenet}
J.~Deng, W.~Dong, R.~Socher, L.~Li, K.~Li, and F.~Li, ``Imagenet: {A}
  large-scale hierarchical image database,'' in \emph{{IEEE CVPR}}.\hskip 1em
  plus 0.5em minus 0.4em\relax {IEEE} Computer Society, 2009, pp. 248--255.

\bibitem{FPGA_survey_Guo2019}
K.~Guo, S.~Zeng, J.~Yu, Y.~Wang, and H.~Yang, ``{[DL] A survey of FPGA-based
  neural network inference accelerators},'' \emph{ACM T RECONFIG TECHN},
  vol.~12, no.~1, pp. 1--26, 2019.

\bibitem{Design-flow-of-accelerating}
J.~Wang, Q.~Lou, X.~Zhang, C.~Zhu, Y.~Lin, and D.~Chen, ``Design flow of
  accelerating hybrid extremely low bit-width neural network in embedded
  fpga,'' in \emph{FPL}.\hskip 1em plus 0.5em minus 0.4em\relax IEEE, 2018, pp.
  163--1636.

\bibitem{winograd}
L.~Lu, Y.~Liang, Q.~Xiao, and S.~Yan, ``Evaluating fast algorithms for
  convolutional neural networks on fpgas,'' in \emph{FCCM}.\hskip 1em plus
  0.5em minus 0.4em\relax IEEE, 2017, pp. 101--108.

\bibitem{caffeine}
C.~Zhang, G.~Sun, Z.~Fang, P.~Zhou, P.~Pan, and J.~Cong, ``Caffeine: Toward
  uniformed representation and acceleration for deep convolutional neural
  networks,'' \emph{IEEE T COMPUT AID D}, vol.~38, no.~11, pp. 2072--2085,
  2018.

\bibitem{Optimizing-FPGA-based-accelerator}
C.~Zhang, P.~Li, G.~Sun, Y.~Guan, B.~Xiao, and J.~Cong, ``Optimizing fpga-based
  accelerator design for deep convolutional neural networks,'' in \emph{FPGA},
  2015, pp. 161--170.

\bibitem{lite-cnn}
M.~V{\'e}stias, R.~P. Duarte, J.~T. de~Sousa, and H.~Neto, ``Lite-cnn: a
  high-performance architecture to execute cnns in low density fpgas,'' in
  \emph{FPL}.\hskip 1em plus 0.5em minus 0.4em\relax IEEE, 2018, pp. 399--3993.

\end{thebibliography}
\vspace{-12mm}
\begin{IEEEbiography}[{\includegraphics[width=1in,height=1.25in,clip]{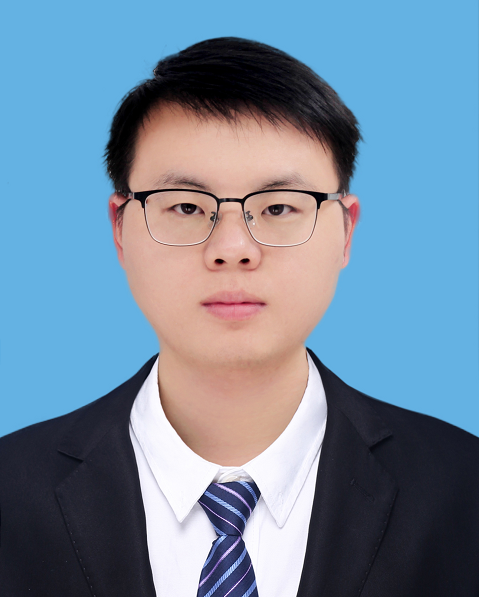}}]{Cheng Gong}
received his B.Eng. degree in computer science from Nankai University in 2016. He is currently working toward his Ph.D. degree in the College of Computer Science, Nankai University. His main research interests include heterogeneous computing, machine learning and Internet of Things.
\end{IEEEbiography}
\vspace{-10mm}
\begin{IEEEbiography}[{\includegraphics[width=1in,height=1.25in,clip]{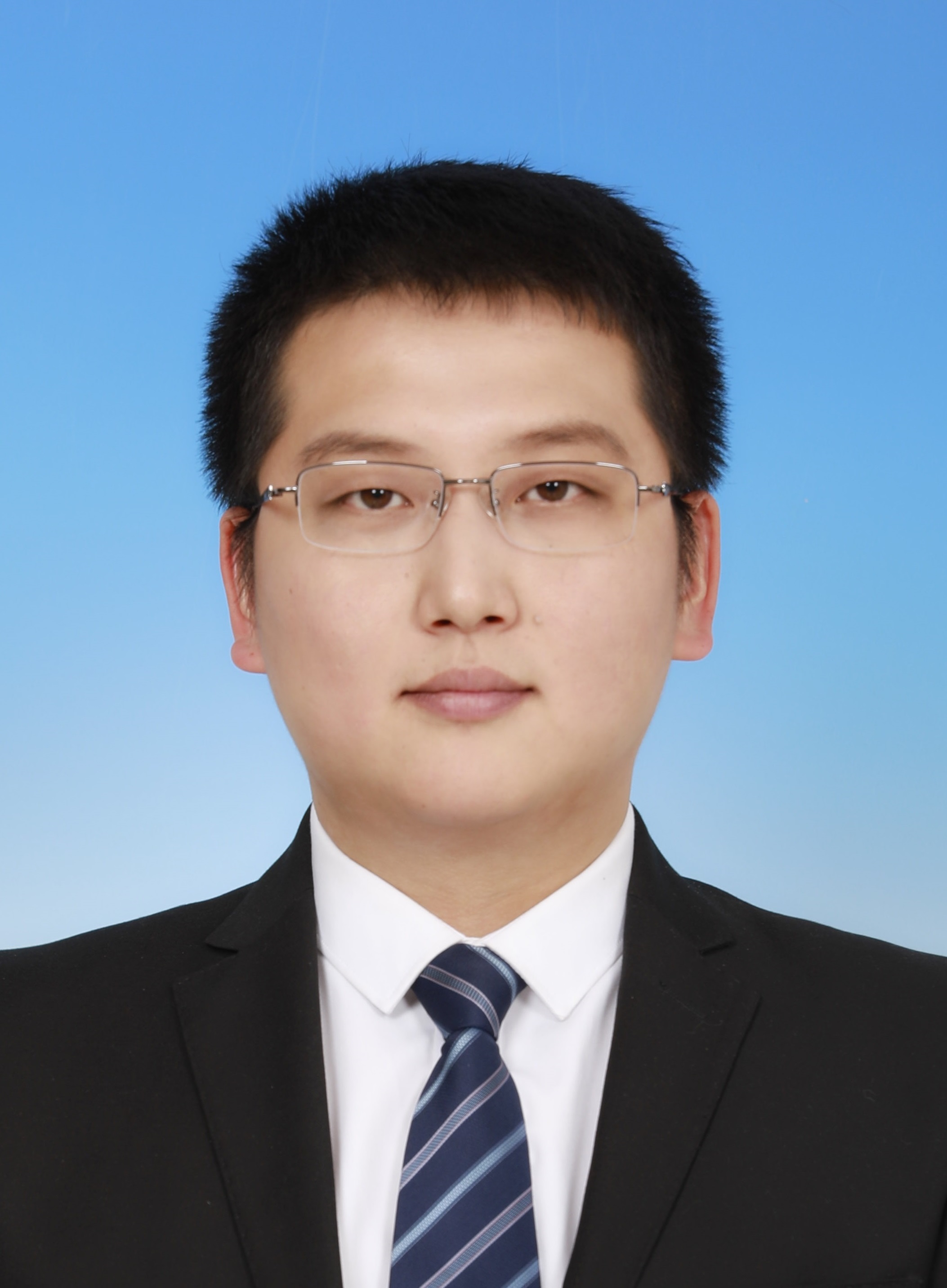}}]{Ye Lu}
received the B.S. and Ph.D. degree from Nankai University, Tianjin, China in 2010 and 2015, respectively. He is an associate professor at the College of Cyber Science, Nankai University now. His main research interests include DNN FPGA accelerator, blockchian virtual machine, embedded system, Internet of Things.
\end{IEEEbiography}
\vspace{-10mm}
\begin{IEEEbiography}[{\includegraphics[width=1in,height=1.25in,clip]{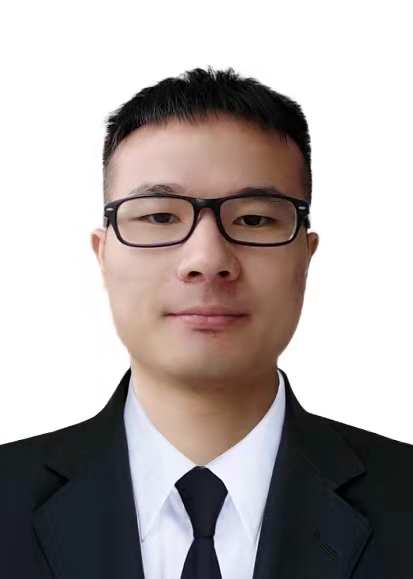}}]{Kunpeng Xie}
received the B.Eng. degree in computer science from Nankai University, in 2019. He is currently working toward the phD in the College of Computer, Intelligent computing system Lab, Nankai University. His research focuses on how computer CPU-GPU heterogeneous system accelerates the training of neural network models and to design effective CNN inference accelerator on FPGA.
\end{IEEEbiography}
\vspace{-10mm}
\begin{IEEEbiography}[{\includegraphics[width=1in,height=1.25in,clip]{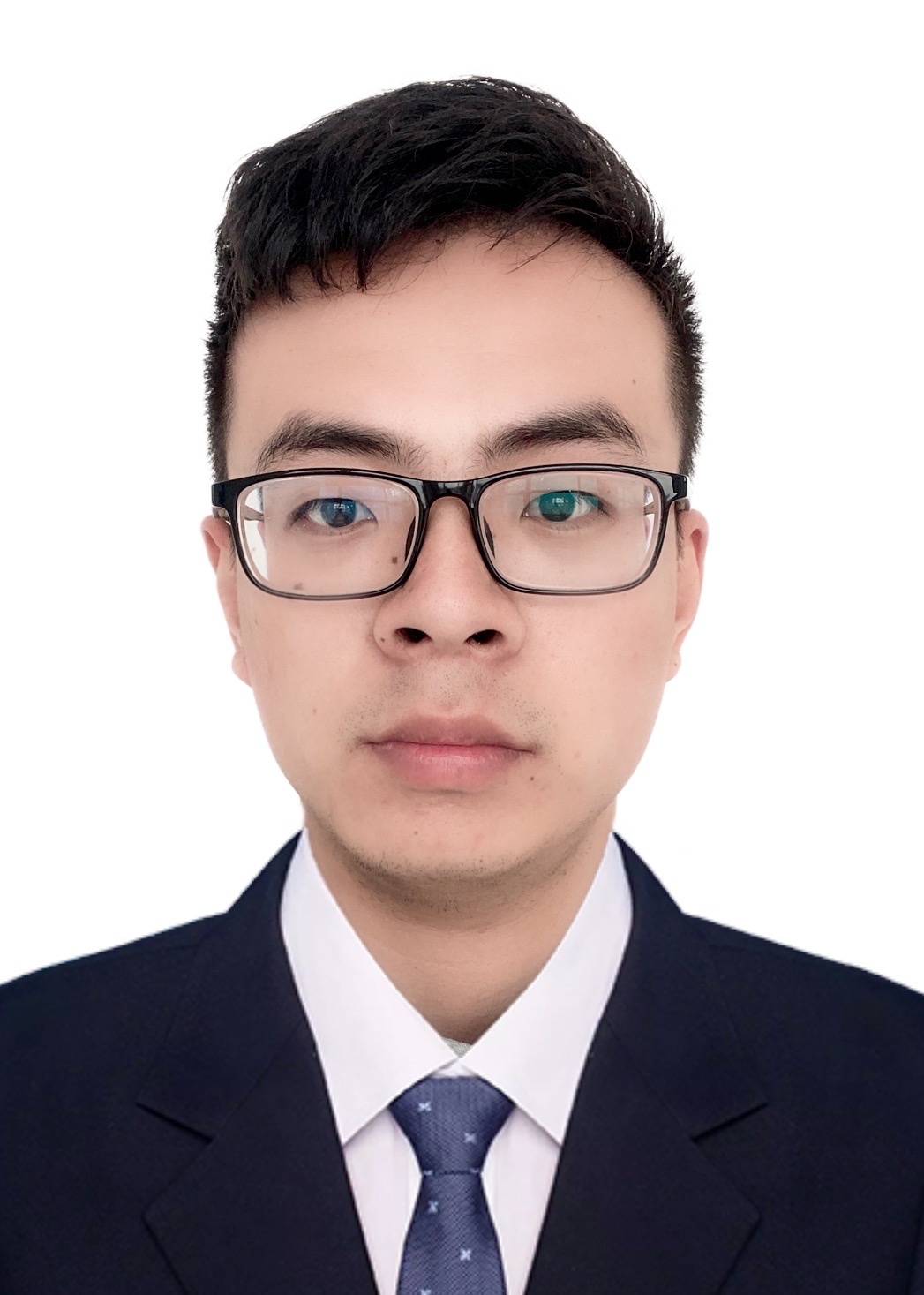}}]{Zongming Jin}
received the B.S. degree from Xidian University, Xian, China, in 2019. He is currently pursuing the M.S. degree at the Intelligent Computing System Lab, College of Cyber Science, Nankai University, Tianjin, China. His Current research interests include heterogeneous computing, embedded system and Internet of things.
\end{IEEEbiography}
\vspace{-10mm}
\begin{IEEEbiography}[{\includegraphics[width=1in,height=1.25in,clip]{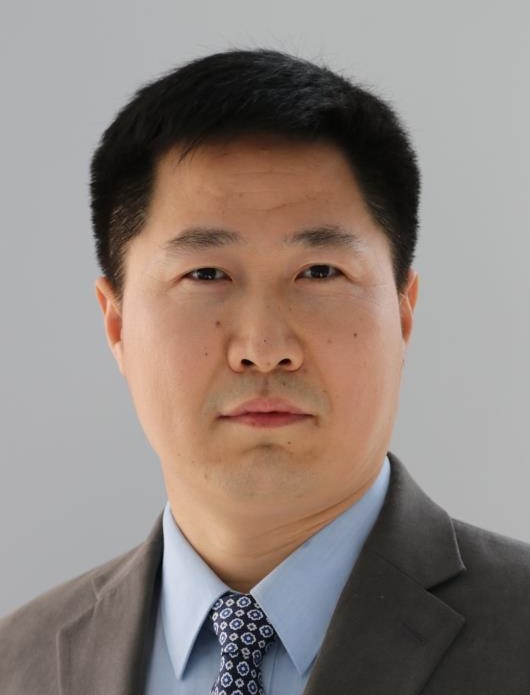}}]{Tao Li}
received his Ph.D. degree in Computer Science from Nankai University, China in 2007. He works at the College of Computer Science, Nankai University as a Professor. He is the Member of the IEEE Computer Society and the ACM, and the distinguished member of the CCF. His main research interests include heterogeneous computing, machine learning and Internet of things.
\end{IEEEbiography}
\vspace{-10mm}
\begin{IEEEbiography}[{\includegraphics[width=1in,height=1.25in,clip]{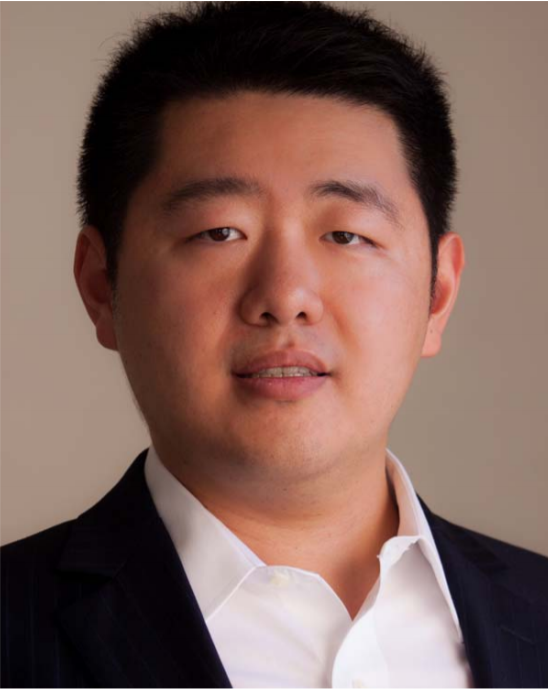}}]{Yanzhi Wang}
received his Ph.D. Degree in Computer Engineering from University of Southern California (USC) in 2014, under the supervision of Prof. Massoud Pedram.
He received the Ming Hsieh Scholar Award (the highest honor in the EE Dept. of USC) for his Ph.D. study. He received his B.S. Degree in Electronic Engineering from Tsinghua University in 2009 with distinction from both the university and Beijing city.
He is currently an assistant professor in the Department of Electrical and Computer Engineering, and Khoury College of Computer Science (Affiliated) at Northeastern University.
His research interests focus on model compression and platform-specific acceleration of deep learning applications.
\end{IEEEbiography}
\end{document}